\pdfoutput=1

\documentclass[11pt,table]{article}
\usepackage[table]{xcolor} 
\usepackage{acl}

\usepackage{times}
\usepackage{latexsym}

\usepackage[T1]{fontenc}

\usepackage[utf8]{inputenc}

\usepackage{microtype}

\usepackage{highlight}
\usepackage{graphicx}
\usepackage{tikz}
\usepackage{comment}
\usepackage{amsmath,amssymb} 
\usepackage{color}
\usepackage{makecell}
\usepackage{multirow}
\usepackage{newfloat}
\usepackage{listings}
\usepackage{array}
\usepackage{subcaption}

\title{Understanding Attention for Vision-and-Language Tasks}
  
\author{Feiqi Cao\textsuperscript{1}, Soyeon Caren Han\textsuperscript{1,2}, Siqu Long\textsuperscript{1}, Changwei Xu\textsuperscript{1}, Josiah Poon\textsuperscript{1}\\
\textsuperscript{1}School of Computer Science, The University of Sydney, Australia\\
\textsuperscript{2}School of Physics, Maths and Computing, The University of Western Australia, Australia\\
\texttt{\{fcao0492, slon6753, chxu7327\}@uni.sydney.edu.au} \\
\texttt{\{caren.han, josiah.poon\}@sydney.edu.au}}

\begin{document}
\maketitle
\begin{abstract}
Attention mechanism has been used as an important component across Vision-and-Language(VL) tasks in order to bridge the semantic gap between visual and textual features. While attention has been widely used in VL tasks, it has not been examined the capability of different attention alignment calculation in bridging the semantic gap between visual and textual clues. In this research, we conduct a comprehensive analysis on understanding the role of attention alignment by looking into the attention score calculation methods and check how it actually represents the visual region's and textual token's significance for the global assessment. We also analyse the conditions which attention score calculation mechanism would be more (or less) interpretable, and which may impact the model performance on three different VL tasks, including visual question answering, text-to-image generation, text-and-image matching (both sentence and image retrieval). Our analysis is the first of its kind and provides useful insights of the importance of each attention alignment score calculation when applied at the training phase of VL tasks, commonly ignored in attention-based cross modal models, and/or pretrained models. Our code is available at: \url{https://github.com/adlnlp/Attention_VL}
\end{abstract}

\section{Introduction}
The relative maturity and flexibility of deep learning allow us to build upon the success of computer vision and natural language processing to face many complex and multimodal Vision-and-Language (VL) tasks, such as Visual Question Answering (VQA), Text-and-Image Matching (T\&I Match), or Text-to-Image Generation (T2I Gen). For these VL tasks, it is crucial to effectively align the multimodal information in both visual and linguistic domains. For example, to pick the right answer in VQA, the model should empower information from the input image, together with aligning the linguistic meanings with visual clues.

\noindent Attention mechanism \citep{bahdanau2015neural, luong-etal-2015-effective} has been used as an important component across a wide range of VL models; from the early-stage attention-based fusion VL models \citep{shih2016look,wang2019camp,xu2018attngan,yang2016stacked}  to the recent VL multimodal transformer-based pretrained models \citep{hu2021vivo,li2020unicoder,lu2019vilbert,ijcai2022p773}. Those attention-based VL models mainly focus on 1) exploring new features to represent visual and linguistic information as an input of attention layer, 2) deciding the position or the number of attentions in the model, or 3) investigating the interpretability of attention distribution on VL tasks by emphasising the specific image regions or textual tokens~\cite{luo2021local}. 

While such approaches and investigations have resulted in interesting findings in different aspects of VL tasks, the attention alignment calculation between vision and language modalities has been less explored. However, the alignment calculation is directly linked to the main purpose of using attention mechanisms in VL tasks, which is to effectively bridge and align two different visual and linguistic information. In other words, the essence of the attention mechanism in VL tasks is the alignment score calculation, as it quantifies the amount of ``Attention’’ that the visual features would place on each of the language representations (or linguistic features would empower on the specific visual regions) when bridging the semantic gap between visual and language features. Most existing VL models directly apply the two attention alignment functions, a general and a dot-product \cite{luong-etal-2015-effective}, which are commonly used in several NLP tasks. Since \citet{vaswani2017transformer} proposed a scaled dot-product for the transformer with full attention, almost every VL paper has directly applied those three attention alignment score functions. Instead, little work has been done towards understanding the role of attention alignment calculation methods applied to bridge visual and linguistic features, and exploring the impact on different VL model performance. 

In order to address this limitation, the overarching goal of this research is to perform an extensive and systematic assessment of the effect of a range of attention alignment mechanisms pertaining to VL tasks, including three major VL tasks: Visual Question Answering (VQA), Text-and-Image Matching (T\&I Match), and Text-to-Image Generation (T2I Gen). Towards that end, we systematically analyse the impact of the position of query and key in attention alignment on VL tasks. We investigate the following three questions: i) Which attention alignment score calculation yields the most benefit in VL tasks? ii) What if we linearly transform the query $Q$ instead of the key $K$ (or vice versa) before the multiplication? For example, assume the textual feature $T$ is a query $Q$, and the image feature $I$ is a key $K$. We analyse the impact of linear transforming $Q$ or $K$ in alignment score calculation. iii) Do the attention alignment calculation techniques with better performance provide better attention distribution interpretability?

In brief, our \textbf{main contributions} are as follows:

\textbf{1)} We conduct a comprehensive analysis of the role of attention alignment score calculation in VL tasks (including three widely-used VL tasks, such as Visual Question Answering, Text-and-Image Matching, and Text-to-Image Generation). \textbf{2)} We perform a comparative analysis of the position of query and key (language and visual feature) for the alignment calculation. \textbf{3)} We evaluate the interpretability of the best and worst attention alignment calculation models. \textbf{4)} We make the code and the data publicly available to encourage reproducibility of results.

\section{Related Works}\label{sec:related}

VL models directly adopt the attention mechanism to bridge the visual and linguistic modal information. In this section, we review the related works for the role of attention mechanisms in different VL tasks within the focus of our analysis.

\textbf{Text-to-Image Generation} AttnGAN \cite{xu2018attngan} first proposed to use dot-product for measuring the alignment between visual subregions and word tokens to guide the image generation process. Many of the later approaches directly adapted the dot-product attention from AttnGAN \citep{han2020victr,li2020manigan,li2019object,pande2021development,qiao2019learn,qiao2019mirrorgan,yin2019semantics,zhu2019dm}. A few models apply the element-wise multiplication \citep{qiao2019learn,qiao2019mirrorgan} or cosine similarity \cite{zhang2021cross} for measuring the attention alignment.

\textbf{Text-and-Image Matching} The cosine similarity based attention alignment proposed by SCAN \cite{lee2018stacked} is most widely used in Text-and-Image Matching \citep{chen2020imram,chen2020expressing,diao2021similarity,dong2021iterative,liu2019focus}. They applied text-to-image (t2i) and image-to-text(i2t) attention in two separate variants to filter the cross-modal relevant representations for later image-sentence matching. Some other approaches applied (scaled) dot-product instead \citep{fei2021cross,liu2020graph,wang2019camp,wei2020multi,Long_2022_WACV}.

\textbf{Visual Question Answering (VQA)} Both textual query-guided image attention and image-guided textual query attention have been commonly used in VQA approaches, which utilised one modality to guide the focus on the other. Several categories of alignment calculations or their variants were included, such as adapting neural networks \citep{anderson2018bottom,patro2018differential,yang2016stacked,zhu2016visual7w} or applying (scaled) dot-product \citep{gao2019dynamic,guo2021re,huang2020aligned,hudson2018compositional,rahman2021improved,yu2019deep,zhang2021dmrfnet} etc. 

\textbf{Text-based Visual Question Answering} Recent TextVQA approaches directly augmented existing VQA models and their cross-modal attention with additional OCR inputs \citep{biten2019icdar, biten2019scene,singh2019towards, wang2020general}. Both early-stage model \textit{M4C} \cite{hu2020iterative} and the most recent pretrained model \textit{TAP} \cite{yang2021tap} fed the question, image and OCR text together into a multimodal transformer and jointly encoded them via scaled-dot product attention in the transformer encoder. 

Nevertheless, there is a lack of research on exploring the most effective cross-modal attention alignment. Hence, we apply different cross-modal attention alignment methods to the most widely adopted baselines for these aforementioned VL downstream tasks: AttnGAN (Text-to-Image Generation), SCAN (Text-to-Image Matching), MAC (VQA), and M4C (TextVQA), and examine the impact of different attention alignment score via in-depth analysis.

\section{Attention Alignment Mechanism} \label{sec:attention}
There are various attention mechanisms applied in different multimodal VL downstream tasks. Two commonly used approaches are the cross-attention and the self-attention. First, the cross-attention is performed between visual and textual inputs. More specifically, given a sequence of textual features $T = \{t_1, t_2, t_3, \ldots, t_M\}$ and image features $I$ = \{$i_1$, $i_2$, $i_3$, \ldots, $i_N$\}, it takes $T$ as the query $Q$ and $I$ as the key $K$ (or vice versa) to compute attention and context vectors $c$ as the attended representations of the input elements in the following way:
\begin{equation}
    a_{xy} = f(Q_x, K_y)
    \label{attention-score}
\end{equation}
\begin{equation}
    \alpha_{xy} = \frac{exp(a_{xy})}{\sum_{y=1}^{n_K} exp(a_{xy})}
    \label{attention-softmax}
\end{equation}
\begin{equation}
    c_{x}^{K} = \sum_{y=1}^{n_K} \alpha_{xy}K_y
    \label{attention-context-vextor}
\end{equation}
where $f$ is a function to calculate attention score, $n_K$ is the number of elements in $K$, and $c_{x}^{K}$ is the context vector of $K$ with respect to the $x$-th element of $Q$. The second approach, self-attention \cite{vaswani2017transformer}, is performed over all inputs from both modalities. In other words, the approach combines $T$ and $I$ as a complete sequence $S = T \cup I$, and converts all elements in $S$ into $Q$, $K$ and $V$ via learnable matrices, which are used to compute attention by multiple heads in the following way:
\begin{equation}
    \mbox{Attention}(Q,K,V) = \mbox{Softmax}(f(Q, K))V
    \label{sa-attention}
\end{equation}

where the results from different heads are combined together. Then, it applies layer normalization, residual connections and fully connected layers in order to obtain the attended representation of the input tokens. With both approaches, we explore the effect of the attention alignment calculation $f$ for different VL tasks with the following five different alignment score functions. We also include \textbf{Cosine similarity}-based attention for only Text-and-Image Matching as it is widely used in that specific domain.

\noindent\textbf{Dot product attention} It was proposed in NMT~\cite{bahdanau2015neural} to compute vector similarity between encoder hidden states and decoder hidden states. This function \cite{luong-etal-2015-effective} has been widely adopted as $f$ in the cross-attention mechanism as shown in Equation \ref{attention-score}.
\begin{equation}
    f(Q, K) = QK
    \label{attention-dot}
\end{equation}

\noindent\textbf{Scaled dot product attention} The higher dimension of data representation would lead to the smaller gradient of softmax function. Hence, the scaling factor was introduced by \cite{vaswani2017transformer}, and applied to the self-attention-based VL approaches as represented in Equation \ref{sa-attention}.
\begin{equation}
    f(Q, K) = \frac{QK}{\sqrt{d}}
    \label{attention-scaled}
\end{equation}

\noindent\textbf{General attention} Along with dot product attention, general attention \cite{luong-etal-2015-effective} received lots of interest as an alternative alignment calculation method that computes attention score using an extra learnable matrix to linearly transform $K$ into the same embedding space as $Q$. This can be considered as one of the neural network based methods mentioned in Section~\ref{sec:related}.
\begin{equation}
    f(Q, K) = QWK
    \label{attention-general}
\end{equation}

\noindent 
There are several variants of neural network based general attention calculation methods. First, \textbf{Biased general attention} is introduced by \cite{Sordoni2016IterativeAN} using more bias towards more important keys regardless of the query context.
\begin{equation}
    f(Q, K) = Q(WK+b)
    \label{attention-biased}
\end{equation}
Secondly, \textbf{Activated general attention.} \cite{ma2017interactive} applies an additional nonlinear activation term, which is able to amplify the emphasis on query elements that are highly relevant to the key. 
\begin{equation}
    f(Q, K) = act(Q(WK+b))
    \label{attention-activated}
\end{equation}
In this paper $act$ is the ReLU activation since it is a widely used function in VL downstream tasks. 

\section{Vision-Language Models}
We use publicly available implementations of the most widely adopted VL baseline models\footnote{All the VL pretrained models are just based on BERT (attention-oriented transformer-based). It is still quite early-stage in this field, and more VL pretrained models are still emerging in 2022~\cite{Zhuge_2021_CVPR,Hong_2021_CVPR}} in order to train and evaluate different attention alignment score calculation for three different VL tasks: (i) \textbf{AttnGan} for Text-to-Image Generation (T2I Gen), (ii) \textbf{SCAN} for Text-and-Image Matching (T\&I Match), (iii) \textbf{MAC} and  \textbf{M4C} for each Visual Question Answering (VQA) and Text-based Visual Question Answering (TVQA).

\subsection{T2I Gen: AttnGAN}
The goal of text-to-image generation is to generate a visually realistic image that matches a given text description. The AttnGAN \cite{Tao18attngan} generates images by using multiple generators with the attention mechanisms. To improve the image quality at each step, a cross-attention mechanism is performed between caption words and image regions, and it produces the attended word context for each image region. Given a caption of $M$ words, an image with $N$ sub-regions would be generated by an upsampling network. The words and image regions are represented as $d$-dimensional vectors \{$t_m$\} $\in T$ and \{$i_n$\} $\in I$ respectively. Then image representation $I$ is applied as $Q$ the query and caption representation $T$ is applied as $K$ the key for the cross-attention mechanism (Equations \ref{attention-score}, \ref{attention-softmax}, \ref{attention-context-vextor}), where the dot product attention score calculation is used as $f$. The resultant textual context would be fused with word region representations as a guide for the generator at the next time step to focus on different words. Note that we evaluate different alignment calculation methods as $f$ to investigate the impact of the image generation performance. We fix $I$ as $Q$ and $T$ as $K$, and replace the dot product with other alignment score calculations.

\subsection{T\&I Match: SCAN}
Text-and-image matching (a.k.a. Text-and-image retrieval) refers to measuring the visual-semantic similarity between a sentence and an image. The SCAN model \cite{lee2018stacked} performs a pair-wise cross-attention between image regions and caption words for fine-grained T\&I Match. This can be done in two directions. Given a caption of $M$ words and an image having $N$ detected objects, $d$-dimensional representations \{$t_m$\} and \{$i_n$\} are obtained as $T$ and $I$ respectively. To obtain the attended image context for each caption word, the cross-attention mechanism (described in Equations \ref{attention-score}, \ref{attention-softmax}) is applied with $T$ being the query $Q$ and $I$ being the key $K$, and an alignment score is measured by using cosine similarity between each caption word and its image context. These alignment scores would be aggregated via a pooling function as the final alignment score between the given image and caption. Such scores can be obtained by using $T$ as $K$ and $I$ as $Q$ to calculate the sentence context for each image region. In experiments, we fix $T$ as $Q$ and $I$ as $K$, and replace the cosine similarity with other alignment score calculations.

\subsection{VQA}
We explore two VQA downstream tasks, Visual Question Answering with compositional reasoning and Text-based Visual Question Answering.   

\subsubsection{VQA: MAC}
First, we focus on the visual question answering task that requires responding to natural language questions about images, specifically with a compositional and structured nature. The MAC circuit \cite{hudson2018compositional} applies a cross-attention mechanism to answer a question based on a given image. Instead of computing attention between textual and visual input, MAC introduces a $d$-dimensional learnable control state $e$ as a guidance for MAC cells to selectively attend to different aspects of inputs at each time step. Within each MAC cell, there is a control unit to attend to the question words and a read unit to attend to the image regions. Given a question of $M$ words and an image having $N$ detected objects, $d$-dimensional representations \{$t_m$\} and \{$i_n$\} are obtained as $T$ and $I$ respectively. Instead of using Equation \ref{attention-score}, the control unit applies $e$ as $Q$ and $T$ as $K$ to compute the attention score in the following way:
\begin{equation}
    a_{y} = W'(f(Q, K_y)) + b'
    \label{attention-score-mac}
\end{equation}
where $f$ indicates element-wise dot product multiplication to obtain a d-dimensional similarity vector, and $W'$ and $b'$ are learnable parameters to output a scalar as the score. Then the control unit follows Equations \ref{attention-softmax}, \ref{attention-context-vextor} to obtain textual context as an update for $e$. Similar to the control unit, the read unit applies $e$ as $Q$ and $I$ as $K$ to obtain the question-guided visual context from the image, which is later aggregated to predict an answer. Therefore the read unit can be considered as a main component in MAC that involves multimodal alignment. Hence, for the evaluation, we fix the control state $e$ (which majorly contains textual question information) as $Q$ and image-based knowledge graph $I$ as $K$, and adapt the focused attention alignment calculation methods $f$ with the element-wise multiplication manner in the read unit.

\subsubsection{TVQA: M4C}
Secondly, Text-based visual question answering (TVQA) is an extension of VQA, which requires the model to read text over the image to answer the questions. The M4C model \cite{hu2020iterative} applies a multimodal transformer over all input modalities to perform iterative answer prediction for the TextVQA task. More specifically, given a question of $M$ words, an image having $N$ detected objects and $O$ detected OCR tokens, $d$-dimensional representations \{$s_m^{ques}$\},  \{$s_n^{obj}$\} and  \{$s_o^{ocr}$\} are obtained as input sequence $S$. The self-attention mechanism (Equation \ref{sa-attention}) with scaled dot product attention is applied over $S$, the sequence of all $M + N + O$ entities. In this way, both intra-modal interactions and inter-modal interactions are captured to aggregate the input to form an answer prediction via classical transformer layers.
Similarly to other tasks, we replace the scaled dot product attention calculation with the other aforementioned options for $f$ to investigate the impact in TVQA.

\section{Evaluation Setup}

\subsection{Datasets}
We conducted experiments on three VL task datasets. The statistics is shown in Table~\ref{tab:split}. We followed the work of the base models, including AttnGAN \cite{Tao18attngan}, SCAN \cite{lee2018stacked}, MAC \cite{hudson2018compositional}, M4C \cite{hu2020iterative} for dataset preprocessing and dividing for train/dev/test.

\begin{table}[t]
    \centering
    \small
    \scalebox{0.68}{
    \begin{tabular}{lcccc}
    \hline
    \textbf{Tasks} & \textbf{Dataset} & \textbf{Train}
    & \textbf{Dev} & \textbf{Test}\\
    \hline
    \multirow{2}{*}{\textbf{T2I Gen}} 
    & CUB & 8,855 & - & 2,933 \\
    & MS-COCO & 82,783 & -  & 15,000 \\
    
    \hline
    \multirow{2}{*}{\textbf{T\&I Match}} 
    & Flickr30k* & 29,000/145,000 & 1,000/5,000 & 1,000/5,000 \\
    & MS-COCO* & 29,000/145,000 &1,000/5,000 & 1,000/5,000\\
    
    \hline
    \multirow{2}{*}{\textbf{VQA}} 
    & CLEVR* & 70,000/699,989 & 15,000/149,991 & 15,000/149,988\\
    & Text-VQA* & 21,953/34,602 & 3,166/5,000 & 3,289/5,734 \\
    \hline
    \end{tabular}}
    \caption{Details of train/dev/test split for each dataset. Note that * indicates the dataset having different numbers for visual and textual inputs. It reports the number of images followed by the number of captions or question-answer pairs, separated by backslash (/).}
    \label{tab:split}
\end{table}

\subsubsection{T2I Gen}
Two benchmark datasets are used: \textbf{Caltech-UCSD Birds 200 (CUB)}\footnote{\url{http://www.vision.caltech.edu/visipedia/CUB-200-2011.html}} and \textbf{MS-COCO}\footnote{\url{https://cocodataset.org/\#home}}. CUB has 11,788 images of 200 bird categories downloaded from the Flickr website, each with 10 textual captions. MS-COCO provides 123,287 images of complex everyday scenes with 5 manually written textual descriptions per image. We use a train/test split of 8,855/2,933 and 82,783/15,000 images respectively for CUB and MS-COCO.

\subsubsection{T\&I Match} \textbf{Flickr30k}\footnote{\url{http://shannon.cs.illinois.edu/DenotationGraph/}} contains around 31k images collected from the Flickr website with 5 crowd-sourced captions per image. We test on Flickr30k with train/dev/test split of 29k/1k/1k images and on MS-COCO (as described above) with 29k/1k/1k images.

\subsubsection{VQA}
We have two VQA tasks: 1) Visual Question Answering with compositional reasoning, and 2) Text-based Visual Question Answering. We used \textbf{CLEVR}\footnote{\url{https://cs.stanford.edu/people/jcjohns/clevr/}} and \textbf{TextVQA}\footnote{\url{https://textvqa.org/dataset}} respectively. CLEVR contains 100,000 synthetic images of 3D shapes with 999,968 questions/answers in total. We use a subset of 70,000 images with 699,989 QAs for training, 15,000 images with 149,991 QAs for validation and 15,000 images with 149,988 QAs for test. TextVQA consists of 45,336 questions asked by (sighted) humans on 28,408 images from the Open Images dataset \cite{krasin2017openimages}. We use the original split: 21,953 images with 35,602 QAs, 3,166 images with 5,000 QAs and 3,289 images with 5,734 QAs for training, validation and test. 

\subsection{Evaluation Metrics}
We describe metrics for assessing the impact of attention alignment mechanism for each VL task.  
\subsubsection{T\&I Match: R@K}
We measure the performance of sentence retrieval and image retrieval by recall at $K$ (R@K), which is defined as the percentage of queries that get the correct item at the closest $K$ points to the query. The higher the value, the better the performance.

\subsubsection{T2I Gen: Inception Score(IS) \& FID}
The evaluation measurement we use is \textbf{Inception Score (IS)} which seeks to capture the image quality and image diversity properties of a collection of generated images. The higher the inception score, the better the model. \textbf{Fr\'echet Inception Distance (FID)} measures the similarity between the generated images and the real images by comparing their Frech\'et distance between the maximum entropy distribution. Lower FID indicates higher similarity.

\begin{table*}[t]
    \centering
    
    \label{tab:results_scan}
    \small
    \scalebox{0.91}{
    \begin{tabular}{lcccccc|cccccc}
    \hline
    \multirow{3}{*}{\textbf{Attention}} & \multicolumn{6}{c|}{\textbf{Sentence Retrieval}} & \multicolumn{6}{c}{\textbf{Image Retrieval}} \\
    & \multicolumn{3}{c}{Flickr30K} & \multicolumn{3}{c|}{MS-COCO} & \multicolumn{3}{c}{Flickr30K} & \multicolumn{3}{c}{MS-COCO} \\
    & R@1 & R@10 & Rsum & R@1 & R@10 & Rsum & R@1 & R@10 & Rsum & R@1 & R@10 & Rsum\\
    \hline
    $\textnormal{cosine similarity}^{\diamond}$
    & \gradientcell{62.4}{55.8}{63.2}{blue}{0.4} 
    & \gradientcell{93.3}{89.7}{93.8}{blue}{0.4} 
    & \gradientcell{243.8}{228.3}{245.0}{blue}{0.4}
    & \gradientcellunderlined{61.4}{52.2}{61.4}{blue}{0.4}
    & \gradientcell{94.5}{91.7}{95.2}{blue}{0.4} 
    & \gradientcell{243}{227.0}{243.7}{blue}{0.4}
    & \gradientcell{43.9}{38.4}{46.7}{blue}{0.4}
    & \gradientcell{81.8}{77.0}{82.1}{blue}{0.4}
    & \gradientcell{199.9}{182.0}{201.8}{blue}{0.4}
    & \gradientcell{45.7}{38.7}{46.0}{blue}{0.4}
    & \gradientcell{88.2}{83.4}{88.3}{blue}{0.4}
    & \gradientcell{212.5}{194.5}{213.4}{blue}{0.4} \\
    
    dot product        
    & \gradientcell{62.1}{55.8}{63.2}{blue}{0.4} 
    & \gradientcell{92.1}{89.7}{93.8}{blue}{0.4} 
    & \gradientcell{240.4}{228.3}{245.0}{blue}{0.4}
    & \gradientcell{59.7}{52.2}{61.4}{blue}{0.4}
    & \gradientcellunderlined{95.2}{91.7}{95.2}{blue}{0.4} 
    & \gradientcell{243.5}{227.0}{243.7}{blue}{0.4}
    & \gradientcell{44.8}{38.4}{46.7}{blue}{0.4}
    & \gradientcellunderlined{82.1}{77.0}{82.1}{blue}{0.4}
    & \gradientcell{200.6}{182.0}{201.8}{blue}{0.4}
    & \gradientcellunderlined{46.0}{38.7}{46.0}{blue}{0.4}
    & \gradientcell{87.9}{83.4}{88.3}{blue}{0.4}
    & \gradientcell{212.6}{194.5}{213.4}{blue}{0.4} \\
    
    scaled dot product 
    & \gradientcell{63.0}{55.8}{63.2}{blue}{0.4}  
    & \gradientcellunderlined{93.8}{89.7}{93.8}{blue}{0.4}
    & \gradientcell{244.9}{228.3}{245.0}{blue}{0.4}
    & \gradientcell{59.3}{52.2}{61.4}{blue}{0.4}
    & \gradientcell{95.1}{91.7}{95.2}{blue}{0.4} 
    & \gradientcellunderlined{243.7}{227.0}{243.7}{blue}{0.4} 
    & \gradientcell{44.9}{38.4}{46.7}{blue}{0.4} 
    & \gradientcell{81.9}{77.0}{82.1}{blue}{0.4}
    & \gradientcell{200.4}{182.0}{201.8}{blue}{0.4}
    & \gradientcell{45.8}{38.7}{46.0}{blue}{0.4}
    & \gradientcellunderlined{88.3}{83.4}{88.3}{blue}{0.4}
    & \gradientcellunderlined{213.4}{194.5}{213.4}{blue}{0.4} \\
    
    general*            
    & \gradientcellunderlined{63.2}{55.8}{63.2}{blue}{0.4} 
    & \gradientcell{93.6}{89.7}{93.8}{blue}{0.4} 
    & \gradientcellunderlined{245.0}{228.3}{245.0}{blue}{0.4}
    & \gradientcell{59.8}{52.2}{61.4}{blue}{0.4} 
    & \gradientcellunderlined{95.2}{91.7}{95.2}{blue}{0.4} 
    & \gradientcellunderlined{243.7}{227.0}{243.7}{blue}{0.4}
    & \gradientcellunderlined{46.7}{38.4}{46.7}{blue}{0.4} 
    & \gradientcell{81.8}{77.0}{82.1}{blue}{0.4}
    & \gradientcellunderlined{201.8}{182.0}{201.8}{blue}{0.4}
    & \gradientcell{45.6}{38.7}{46.0}{blue}{0.4}
    & \gradientcell{87.8}{83.4}{88.3}{blue}{0.4}
    & \gradientcell{212.1}{194.5}{213.4}{blue}{0.4} \\
    
    $\textnormal{general}^{\dagger}$   
    & \gradientcell{56.6}{55.8}{63.2}{blue}{0.4} 
    & \gradientcell{90.1}{89.7}{93.8}{blue}{0.4} 
    & \gradientcell{229.9}{228.3}{245.0}{blue}{0.4}
    & \gradientcell{53.8}{52.2}{61.4}{blue}{0.4}
    & \gradientcell{93.1}{91.7}{95.2}{blue}{0.4} 
    & \gradientcell{231.5}{227.0}{243.7}{blue}{0.4}
    & \gradientcell{38.4}{38.4}{46.7}{blue}{0.4} 
    & \gradientcell{77.0}{77.0}{82.1}{blue}{0.4}
    & \gradientcell{182.0}{182.0}{201.8}{blue}{0.4}
    & \gradientcell{39.3}{38.7}{46.0}{blue}{0.4}
    & \gradientcell{83.4}{83.4}{88.3}{blue}{0.4}
    & \gradientcell{195.3}{194.5}{213.4}{blue}{0.4} \\
    
    biased general*     
    & \gradientcell{56.6}{55.8}{63.2}{blue}{0.4} 
    & \gradientcell{89.8}{89.7}{93.8}{blue}{0.4} 
    & \gradientcell{230.3}{228.3}{245.0}{blue}{0.4}
    & \gradientcell{52.2}{52.2}{61.4}{blue}{0.4} 
    & \gradientcell{91.7}{91.7}{95.2}{blue}{0.4} 
    & \gradientcell{227.0}{227.0}{243.7}{blue}{0.4}
    & \gradientcell{39.6}{38.4}{46.7}{blue}{0.4} 
    & \gradientcell{77.3}{77.0}{82.1}{blue}{0.4}
    & \gradientcell{185.0}{182.0}{201.8}{blue}{0.4}
    & \gradientcell{38.7}{38.7}{46.0}{blue}{0.4} 
    & \gradientcell{83.4}{83.4}{88.3}{blue}{0.4}
    & \gradientcell{194.5}{194.5}{213.4}{blue}{0.4} \\
    
    $\textnormal{biased general}^{\dagger}$
    & \gradientcell{55.8}{55.8}{63.2}{blue}{0.4} 
    & \gradientcell{89.7}{89.7}{93.8}{blue}{0.4} 
    & \gradientcell{228.3}{228.3}{245.0}{blue}{0.4} 
    & \gradientcell{52.6}{52.2}{61.4}{blue}{0.4} 
    & \gradientcell{93.2}{91.7}{95.2}{blue}{0.4} 
    & \gradientcell{231.1}{227.0}{243.7}{blue}{0.4}
    & \gradientcell{39.3}{38.4}{46.7}{blue}{0.4} 
    & \gradientcell{77.4}{77.0}{82.1}{blue}{0.4}
    & \gradientcell{184.6}{182.0}{201.8}{blue}{0.4}
    & \gradientcell{39.8}{38.7}{46.0}{blue}{0.4}
    & \gradientcell{84.2}{83.4}{88.3}{blue}{0.4}
    & \gradientcell{197.3}{194.5}{213.4}{blue}{0.4} \\
    
    activated general  
    & \gradientcell{56.2}{55.8}{63.2}{blue}{0.4} 
    & \gradientcell{90.5}{89.7}{93.8}{blue}{0.4} 
    & \gradientcell{229.2}{228.3}{245.0}{blue}{0.4}
    & \gradientcell{53.9}{52.2}{61.4}{blue}{0.4}
    & \gradientcell{92.9}{91.7}{95.2}{blue}{0.4} 
    & \gradientcell{231.3}{227.0}{243.7}{blue}{0.4}
    & \gradientcell{39.2}{38.4}{46.7}{blue}{0.4} 
    & \gradientcell{77.4}{77.0}{82.1}{blue}{0.4}
    & \gradientcell{184.7}{182.0}{201.8}{blue}{0.4}
    & \gradientcell{39.5}{38.7}{46.0}{blue}{0.4}
    & \gradientcell{84.0}{83.4}{88.3}{blue}{0.4}
    & \gradientcell{195.6}{194.5}{213.4}{blue}{0.4} \\
    
    \hline
    \end{tabular}}
    \caption{R@1, R@10 and the sum of (R@1+R@5+R@10) on Flickr30K and MS-COCO for T\&I Match. The definition of $\diamond$, *, $\dagger$ can be found in footnote 8. $Q$ refers to caption words and $K$ refers to image regions.}
    \label{tab:results_scan}
\end{table*}

\subsubsection{VQA}
For the \textbf{VQA with compositional reasoning}, overall accuracy is used to measure the performance of the VQA models. The higher the accuracy, the better the performance of the model. For the \textbf{TVQA}, it is designed for the VQA context where 10 ground truth answers are provided for each question-image pair. The accuracy of a single prediction is a soft score obtained by a vote of the 10 ground truth answers. Overall accuracy is obtained by taking the average across all instances. We also use the Average Normalized Levenshtein Similarity (ANLS) score \cite{biten2019scene} to eliminate the dropped performance caused by OCR recognition error by comparing the string similarity between the ground truth and the prediction.

\subsection{Experimental Settings}\label{sec:appendix-setup}
For \textbf{T\&I Match: SCAN} (t-i) AVG models, all settings of hyper-parameters follow the configuration of the SCAN. The batch size is 128, the margin of triplet loss $\alpha$ is 0.2 and the threshold of maximum gradient norm for gradient clipping is 2. For Flickr30k models, the learning rate is set as 0.0002 for the first 15 epochs and then lowered to 0.00002 for another 15 epochs. Total training epochs are 30 and the best model is selected with the highest sum of R@K score. For MS-COCO models, we trained with a learning rate of 0.0005 for 10 epochs and then lowered the learning rate to 0.00005. The best model is selected with the highest sum of R@K score. Training epochs are 20. For \textbf{T2I Gen: AttnGan} model on CUB dataset, the batch size is set to be 20 and we trained with 400 epochs in total. On the MS-COCO dataset, the batch size is 14 and total epochs are 90. In addition to this, all settings are the same as the AttnGan. For \textbf{VQA: MAC} models, the training epoch is set to be 8 and other hyperparameter settings are consistent with MAC. More specifically, the batch size is 128, the learning rate is 0.0001 with 0.5 learning decay rate and the threshold of maximum gradient norm for gradient clipping is 8. For \textbf{TVQA: M4C} model on the Text-VQA dataset, we followed the exact same setting as M4C, applying the batch size of 128 and 100 epochs for training, All model variants would train to convergence within 80 epochs.

All experiments for T2I Gen, T\&I Match and VQA are conducted on a variety of cloud instances from Google Colab, with each utilising an NVIDIA Tesla T4 GPU of 16GB RAM.
For TVQA the experiments are conducted utilising NVIDIA Titan RTX GPU with 24GB RAM, 16 Intel(R) Core(TM) i9-9900X CPU @ 3.50GHz with 128GB RAM, and the operating system of Ubuntu 20.04.1.


\section{Results}
We analyse the impact of attention alignment mechanisms in different VL tasks, and explore the interpretability based on attention distribution.

\subsection{Test Performance}
A primary goal of this work is to identify the most effective and successful attention alignment calculation functions for VL tasks. Tables \ref{tab:results_scan}, \ref{tab:results_vqa}, and \ref{tab:results_t2i} \footnote{$\diamond$ indicates the original attention alignment function used by the base models. * indicates $f(K,Q)$ (swapping query and key), and $\dagger$ indicates $f(Q,K)$ (without swapping query and key) for Equations \ref{attention-general} and \ref{attention-biased}} detail the results of our experiments comparing performance of individual alignment functions with each VL models. Each table visualises the trends with a heatmap. The darker the colour of the cells, the better the performance. As shown in Table \ref{tab:results_scan} for the T\&I Match task that the original calculation function, cosine similarity, achieved quite good performance. However, scaled dot product and general* demonstrated a consistent superiority for both sentence retrieval and image retrieval on both Flickr30K and MS-COCO. Comparatively, biased and activated general attentions produced very low results overall.

\begin{table}[t]
    \centering
    \footnotesize
    
    \begin{minipage}[t]{0.4\linewidth}\centering
        \scalebox{0.9}{
        \setlength\tabcolsep{2pt}
        \begin{tabular}{lc}
        \hline
        \textbf{Attention} & \textbf{Acc.}\\
        \hline
        $\textnormal{dot product}^{\diamond}$                     
        &\gradientcell{0.966}{0.9592}{0.9734}{blue}{0.45}\\
        
        scaled dot product                
        &\gradientcellunderlined{0.973}{0.9592}{0.9734}{blue}{0.45}\\
        
        general*
        &\gradientcell{0.967}{0.9592}{0.9734}{blue}{0.7}\\
        
        $\textnormal{general}^{\dagger}$
        &\gradientcell{0.962}{0.9592}{0.9734}{blue}{0.7}\\
        
        biased general* 
        &\gradientcell{0.959}{0.9592}{0.9734}{blue}{0.25}\\
        
        $\textnormal{biased general}^{\dagger}$
        &\gradientcell{0.963}{0.9592}{0.9734}{blue}{0.25}\\
        
        activated general
        &\gradientcell{0.971}{0.9592}{0.9734}{blue}{0.45}\\
        \hline
        \\
        \multicolumn{2}{c}{\footnotesize (a) VQA on CLEVR} \\
        \end{tabular}}
    \end{minipage}\hfill
    \begin{minipage}[t]{0.6\linewidth}\centering
        \scalebox{0.9}{
        \setlength\tabcolsep{2pt}
        \begin{tabular}{lcc}
        \hline
        \textbf{Attention} & \textbf{Acc.} & \textbf{ANLS} \\
        \hline
        dot product                      
        &\gradientcell{0.407}{0.4065}{0.4189}{blue}{0.45}
        &\gradientcell{0.545}{0.5445}{0.5544}{blue}{0.45}\\
        
        $\textnormal{scaled dot product}^{\diamond}$             
        & \gradientcellunderlined{0.419}{0.4065}{0.419}{blue}{0.45}
        & \gradientcellunderlined{0.554}{0.5445}{0.554}{blue}{0.45}\\
        
        general*
        & \gradientcell{0.407}{0.4065}{0.4189}{blue}{0.9}
        & \gradientcell{0.546}{0.5445}{0.5544}{blue}{0.7}\\
        
        $\textnormal{general}^{\dagger}$
        & \gradientcell{0.416}{0.4065}{0.4189}{blue}{0.55}
        & \gradientcellunderlined{0.554}{0.5445}{0.5544}{blue}{0.5}\\
        
        biased general* 
        & \gradientcell{0.412}{0.4065}{0.4189}{blue}{0.25}
        & \gradientcell{0.553}{0.5445}{0.5544}{blue}{0.25}\\
        
        $\textnormal{biased general}^{\dagger}$
        & \gradientcell{0.414}{0.4065}{0.4189}{blue}{0.25}
        & \gradientcell{0.551}{0.5445}{0.5544}{blue}{0.25}\\
        
        activated general
        & \gradientcell{0.413}{0.4065}{0.4189}{blue}{0.25}
        & \gradientcell{0.548}{0.5445}{0.5544}{blue}{0.25}\\
        \hline
        \\
        \multicolumn{3}{c}{\footnotesize (b) TVQA on Text-VQA} \\
        \end{tabular}}
    \end{minipage}
    \caption{Results for VQA/TVQA. The definitions of $\diamond$, *, $\dagger$ are in footnote 8. For VQA, $Q$ refers to the control state and $K$ refers to image-based knowledge graph in read unit. For TVQA, $Q$ and $K$ are transformed union of all caption words, image object and OCR features.}
    \label{tab:results_vqa}
\end{table}

Table \ref{tab:results_vqa} details the performance of alignment functions for VQA tasks, including (a) VQA with compositional reasoning (CLEVR) and (b) TVQA (Text-VQA). Surprisingly, both VQA and TVQA models produced the best performance with a scaled dot product alignment, highlighting its overall effectiveness for the VQA tasks. We note that the activated general attention (ReLU activation) performed well for VQA but produced one of the lowest ANLS scores in TVQA. The general attention alignment function also showed the similar trend. Considering the different nature of general VQA and TVQA, where the latter mainly focuses on OCR text input, it is unsurprising that the impact of alignment mechanism is discrepant. Hence, it is remarkable to find that the scaled dot product achieved the best in both tasks.

\begin{table}[t!]
    \centering
    \footnotesize
    \scalebox{1}{
    \begin{tabular}{lcccc}
    \hline
    \multirow{2}{*}{\textbf{Attention}} 
    & \multicolumn{2}{c}{CUB} & \multicolumn{2}{c}{MS-COCO}\\
    & IS & FID & IS & FID \\
    \hline
    $\textnormal{dot product}^{\diamond}$                      
    & \gradientcell{4.32}{4.13}{4.41}{blue}{0.4}
    & \gradientcellunderlined{25.72}{28.39}{25.72}{blue}{0.4}
    & \gradientcell{23.28}{23.05}{25.24}{blue}{0.4}
    & \gradientcellunderlined{40.19}{43.64}{40.19}{blue}{0.4}\\
    
    scaled dot product                
    & \gradientcell{4.31}{4.13}{4.41}{blue}{0.4}
    & \gradientcell{25.74}{28.39}{25.72}{blue}{0.4}
    & \gradientcell{23.84}{23.05}{25.24}{blue}{0.4}
    & \gradientcell{42.33}{43.64}{40.19}{blue}{0.4}\\
    
    general*
    & \gradientcell{4.36}{4.13}{4.41}{blue}{0.4}
    & \gradientcell{28.21}{28.39}{25.72}{blue}{0.4}
    & \gradientcell{24.28}{23.05}{25.24}{blue}{0.4}
    & \gradientcell{40.82}{43.64}{40.19}{blue}{0.4}\\
    
    $\textnormal{general}^{\dagger}$ 
    & \gradientcell{4.26}{4.13}{4.41}{blue}{0.4}
    & \gradientcell{26.94}{28.39}{25.72}{blue}{0.4}
    & \gradientcell{24.63}{23.05}{25.24}{blue}{0.4}
    & \gradientcell{42.45}{43.64}{40.19}{blue}{0.4}\\
    
    biased general* 
    & \gradientcell{4.13}{4.13}{4.41}{blue}{0.4}
    & \gradientcell{26.97}{28.39}{25.72}{blue}{0.4}
    & \gradientcell{23.05}{23.05}{25.24}{blue}{0.4}
    & \gradientcell{43.10}{43.64}{40.19}{blue}{0.4}\\
    
    $\textnormal{biased general}^{\dagger}$ 
    & \gradientcell{4.30}{4.13}{4.41}{blue}{0.4}
    & \gradientcell{25.89}{28.39}{25.72}{blue}{0.4}
    & \gradientcellunderlined{25.24}{23.05}{25.24}{blue}{0.4}
    & \gradientcell{43.64}{43.64}{40.19}{blue}{0.4}\\
    
    activated general
    & \gradientcellunderlined{4.41}{4.13}{4.41}{blue}{0.4}
    & \gradientcell{28.39}{28.39}{25.72}{blue}{0.4}
    & \gradientcell{23.56}{23.05}{25.24}{blue}{0.4}
    & \gradientcell{42.65}{43.64}{40.19}{blue}{0.4}\\
    \hline
    
    \end{tabular}}
    \caption{Results on CUB and MS-COCO for T2I Gen. The definitions of $\diamond$, *, $\dagger$ are in footnote 8. $Q$ refers to caption words and $K$ refers to image subregions.}
    \label{tab:results_t2i}
\end{table}

The T2I Gen results in Table \ref{tab:results_t2i} illustrated quite different trends compared to the two aforementioned tasks. First, none of the alignment functions produced a consistently better performance in both evaluation metrics, IS or FID. While neural network-based alignment functions (i.e. general, biased and activated general) achieved higher IS scores than others, the dot product dominated in FID scores, for both CUB and MS-COCO. The scaled dot product obtained comparably good FID results but not in IS. This can be aligned with the different metrics of measurement that FID counts on the similarity between the ground-truth images and the images generated from the text whereas IS expects the diversity of the generated image. Hence, a better VL alignment leads to the better FID but not necessarily the better IS.

In summary, we can find the scaled dot product can be the best alignment calculation function for both cross and self-attention that can successfully bridge the visual and textual information, as it produces considerably and consistently better results for all three VL tasks across all six VL datasets.

\subsection{Impact of Key and Query}\label{sec:kq-impact}
Since the previous works~\cite{luong2015effective,Sordoni2016IterativeAN,ma2017interactive} do not have a standard choice of linearly transforming key or value when calculating attention scores, we also investigated the impact of position of query and key in the attention alignment calculation process, especially when extra learnable weights and biases are involved. We explore the difference between linearly transforming the key $K$ to multiply with the query $Q$ ($f(K,Q) = KWQ$) and transforming the query $Q$ to multiply with the key $K$  ($f(Q,K) = QWK$) in general attention and biased general attention calculation. Specifically, we initially fixed the textual information as a query $Q$ and visual information as a key $K$ (Equation \ref{attention-general} and \ref{attention-biased}) and swapped the position in different general attention alignment score measurements. In Table \ref{tab:results_scan}, \ref{tab:results_vqa}, and \ref{tab:results_t2i}, * indicates the functions with $f(K,Q)$, whereas  $\dagger$ refers to those with  $f(Q,K)$.

Table \ref{tab:results_scan} shows that Flickr30K performed better with general* or biased general*, whereas MS-COCO does not have obvious trends. Similar patterns can be found in both cross-attention mechanisms (VQA models, T2I Gen models), and self-attention mechanism TVQA models. Interestingly but unsurprisingly, we note that there is no obvious and consistent performance improvement pattern in different positions of textual information (query $Q$) and visual information (key $K$) when it calculates the alignment. It depends on the specific downstream tasks and dataset. We can conclude that the way of calculating alignment is the crucial point in VL tasks, compared to the position/order of different modal information.

\begin{table}[t]
    \centering
    \scalebox{0.6}{
    \begin{tabular}{lccccc}
    \hline
    \multirow{3}{*}{\textbf{Attention}} &
    \multirow{2}{*}{\textbf{Exist}} &
    \textbf{Query} &
    \textbf{Compare} &
    \multirow{2}{*}{\textbf{Count}} & 
    \textbf{Compare} \\
    & & \textbf{Attribute}
    &\textbf{Attribute} 
    &
    & \textbf{Integer}\\
    & (single) 
    & (single)
    & (two) 
    & (multiple)
    & (multiple)\\    
    \hline

    scaled dot product
    
    & \gradientcellunderlined{0.9912}{0.8938}{0.9928}{blue}{0.1}
    & \gradientcellunderlined{0.9928}{0.8938}{0.9928}{blue}{0.1}
    & \gradientcellunderlined{0.9860}{0.8938}{0.9928}{blue}{0.1}
    & \gradientcellunderlined{0.9200}{0.8938}{0.9928}{blue}{0.9}
    & \gradientcellunderlined{0.9780}{0.8938}{0.9928}{blue}{0.4}\\

    biased general*     
    & \gradientcell{0.9883}{0.8938}{0.9928}{blue}{0.1}
    & \gradientcell{0.9920}{0.8938}{0.9928}{blue}{0.1}
    & \gradientcell{0.9832}{0.8938}{0.9928}{blue}{0.1}
    & \gradientcell{0.8938}{0.8938}{0.9928}{blue}{0.4}
    & \gradientcell{0.9014}{0.8938}{0.9928}{blue}{0.4}\\
    \hline
    \end{tabular}}
    \caption{Breakdown analysis of VQA accuracy regarding different question types. The number of queried objects in the questions are included in the brackets.}
    \label{tab:results_breakdown_vqa}
\end{table}

\subsection{Breakdown Analysis} \label{sec:breakdown}

\paragraph{VQA} To further investigate the difference between attention calculation methods, we report the accuracy for the best/worst performing attention calculation methods regarding different question categories in CLEVR. As shown in Table~\ref{tab:results_breakdown_vqa}, the two models achieved similar performance for question types \textit{Exist}, \textit{Query Attribute}, and \textit{Compare Attribute}. \textit{Exist} and \textit{Query Attribute} types normally contain single queried object in the question, and \textit{Compare Attribute} questions would contain two queried objects. Those question types only require models to attend to one or two objects in the image, so it is easier for model to capture the pattern when learning the alignment between image and question. However, the \textit{Count} and \textit{Compare Integer} questions are challenging to answer as finding multiple objects with the same attributes is required. The models need to learn how to align multiple objects to one noun phrase. In this case, scaled dot product attention always works better than biased general attention by up to 7.66\% accuracy, which suggests that scaled dot product attention can learn more accurate alignment between image regions and question words.

\begin{figure}[t!]
\centering
  \begin{subfigure}[t]{0.95\columnwidth}
    \includegraphics[width=\linewidth]{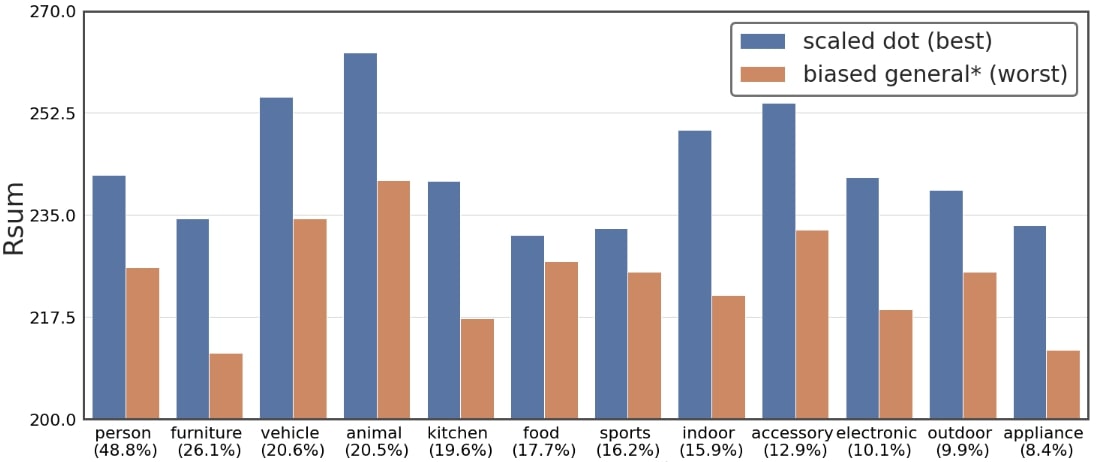}
    \caption{Text Retrieval}
    \label{fig:TR_breakdown_supercat}
  \end{subfigure}
  \hfill
  \begin{subfigure}[t]{0.95\columnwidth}
    \includegraphics[width=\linewidth]{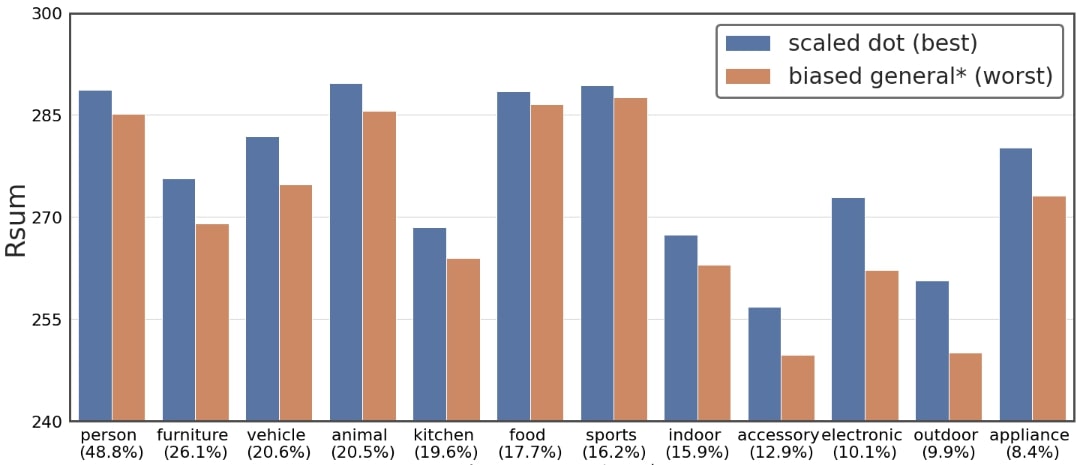}
    \caption{Image Retrieval}
    \label{fig:IR_breakdown_supercat}
  \end{subfigure}

  \caption{Breakdown analysis of T\&I Match on MSCOCO image supercategories.} \label{fig:breakdown_supercat}
\end{figure}

\begin{figure}[t!]
\centering
  \begin{subfigure}[t]{0.47\columnwidth}
    \includegraphics[width=\linewidth]{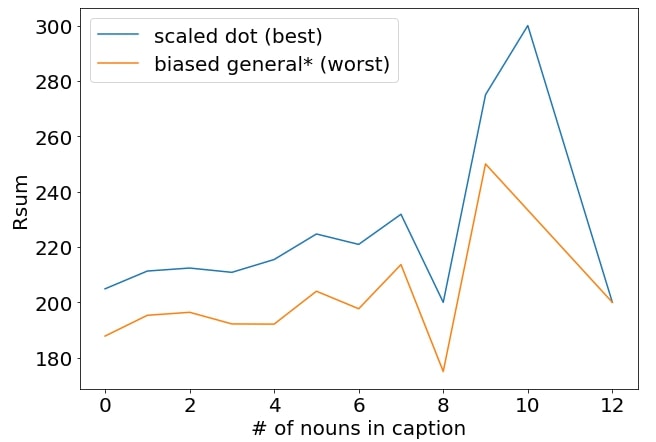}
    \caption{Image Retrieval}
    \label{fig:IR_breakdown_nouns}
  \end{subfigure}
  \begin{subfigure}[t]{0.47\columnwidth}
    \includegraphics[width=\linewidth]{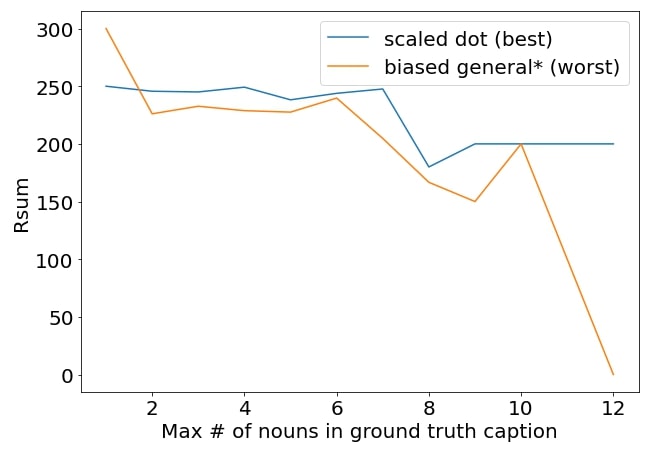}
    \caption{Text Retrieval}
    \label{fig:TR_breakdown_nouns}
  \end{subfigure}
  \caption{Breakdown analysis of T\&I Match on the number of nouns in MS-COCO captions.} \label{fig:breakdown_nouns}
\end{figure}

\begin{figure}[t]
    \centering
    \includegraphics[width=0.97\linewidth]{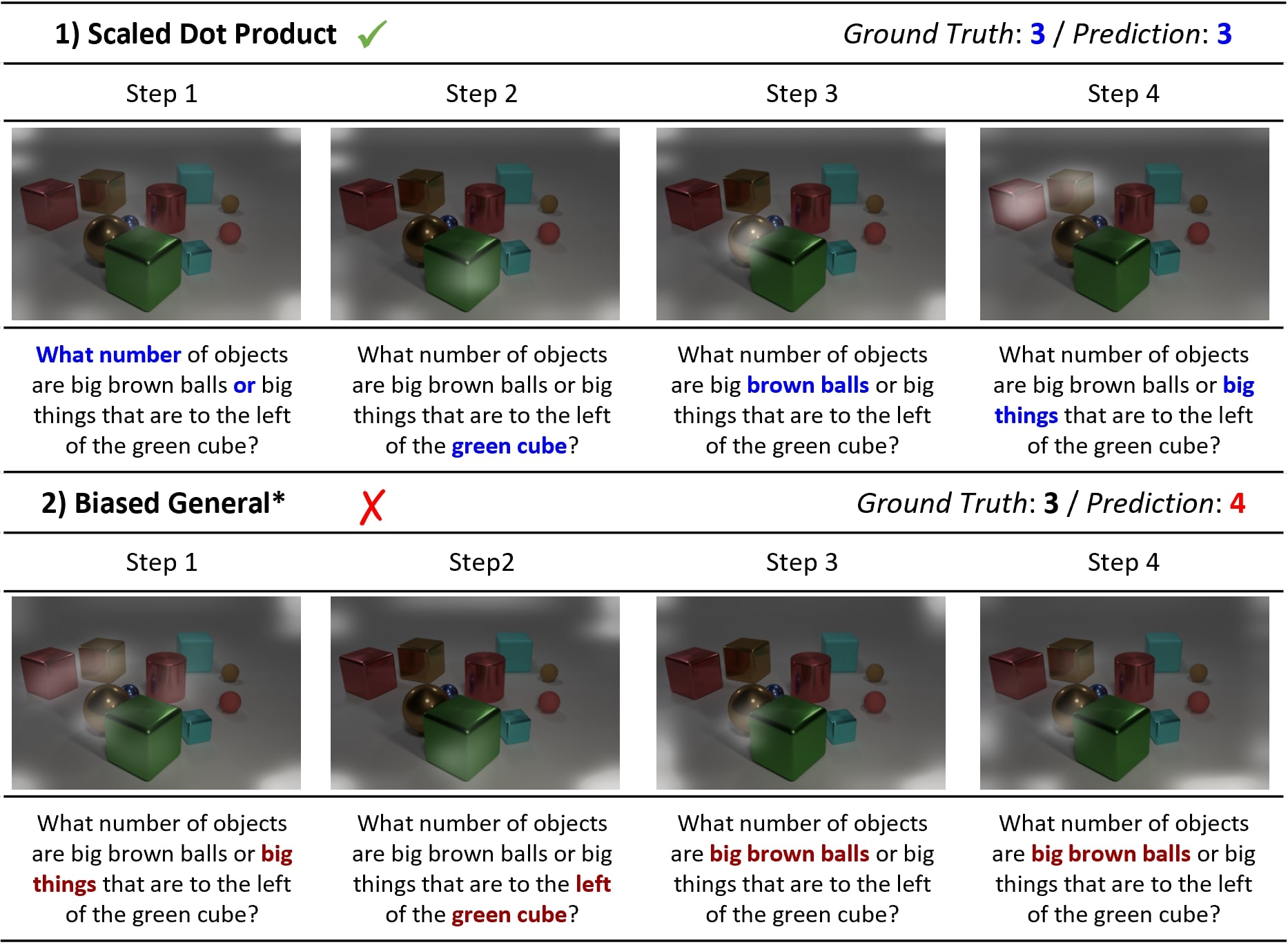}
    \caption{Qualitative example of VQA-CLEVR from the MAC trained by different attention alignment.}
    \label{fig:mac-analysis-appendix-1}
\end{figure}

\paragraph{T\&I Match} The VQA (CLEVR) only includes a limited set of objects with limited attribute descriptions, so we also investigated the effect of real-world images with more diverse types of objects and descriptions using MS-COCO. We compared the retrieval Rsum of the best performing and the worst performing attention calculation methods in terms of different image annotation supercategories in Figure~\ref{fig:breakdown_supercat} and the number of caption nouns in Figure~\ref{fig:breakdown_nouns}. The supercategories on the x-axis in Figure~\ref{fig:breakdown_supercat} are sorted based on the percentage of the test set images which contain that specific supercategory in the annotations, as indicated by the value in the brackets below each supercategory. From Figure~\ref{fig:TR_breakdown_supercat}, it is clearly observed that scaled dot product attention can consistently perform much better than the biased general* attention for most categories such as \textit{person}, \textit{vehicle} and \textit{electronic}, which are easier to be distinguished based on consistent visual features or shapes. Visual elements under \textit{food} and \textit{sports} categories are difficult to be distinguished and aligned due to vastly different types of visual cues (such as shapes, colors and textures). However, for image with annotations of these challenging categories, scaled dot product still manages to outperform biased general* attention, even though by a smaller performance gain. Regarding the image retrieval task, we can observe a similar pattern. Visual element like \textit{electronic} and \textit{outdoor} can have consistent and common linguistic terms such as phones, microwaves, fridges and fields that can be easily distinguished in the description, therefore those images are easier to be aligned to and retrieved given a text. In this case, scaled dot product can outperform biased general* attention by a large margin. However, for other image supercategories such as \textit{food}, \textit{sports} and \textit{person}, the description can be very different due to subjectivity and variety of phrasing choices, making them more difficult to be aligned to and retrieved. Therefore the scaled dot product attention only manages to perform better than biased general* attention by a relatively small margin. Based on the observations above, we can conclude that the scaled dot product can perform better than biased general* attention in visual-linguistic alignment, especially for easily alignable linguistic and visual cues.


In terms of the number of nouns to be aligned in the captions, we can see from Figure~\ref{fig:IR_breakdown_nouns} that scaled dot product attention can maintain a consistent performance margin over biased general* attention when retrieving images, regardless of whether there is no object, only a few objects or many objects in the caption to be aligned. However, they performed the same at 12 as there is only one caption with 12 nouns in the test set and the models cannot really be distinguished on the single instance. When we group the images based on the maximum number of objects that can be possibly contained in their descriptions, we can see from Figure~\ref{fig:TR_breakdown_nouns} that scaled dot product attention can still outperform biased general* attention in most cases when retrieving relevant descriptions except for the two image query instances with the maximum of only one possible object to be aligned in the description. Based on the above patterns, we can conclude that scaled dot product can generally learn better visual-linguistic alignment than biased general* attention.

\subsection{Qualitative Analysis}
\begin{figure}[t]
    \centering
    \includegraphics[width=0.95\linewidth]{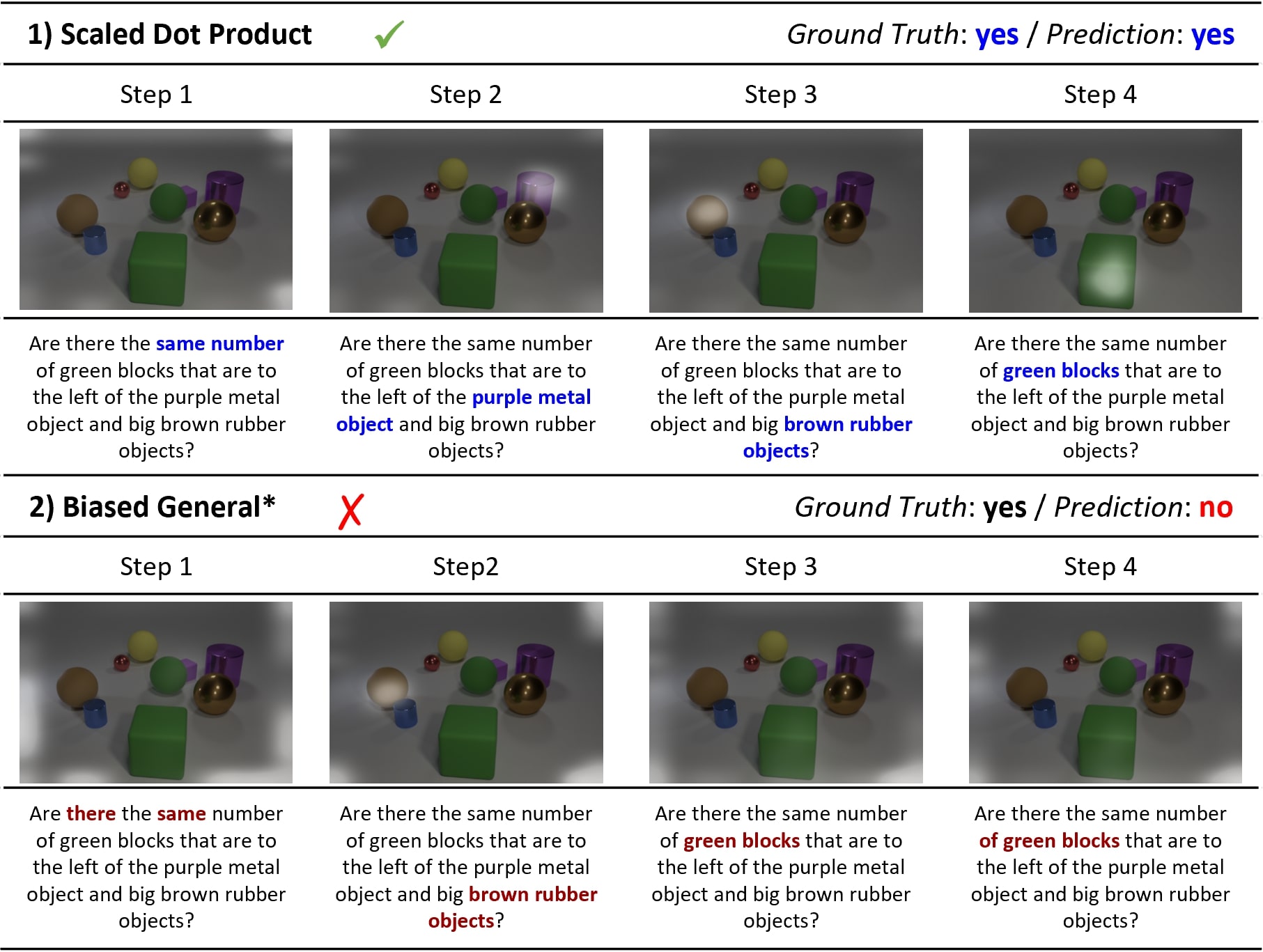}
    \caption{Qualitative example of VQA-CLEVR from the MAC trained by different attention functions.}
    \label{fig:mac-analysis-appendix-2}
\end{figure}

We visualised the prediction interpretability of the best and worst attention alignment calculation for VQA task on the question category \textit{Count} and \textit{Compare Integer}. More examples on other tasks can be found in Appendix~\ref{sec:qualitative-appendix}. Figure~\ref{fig:mac-analysis-appendix-1} shows a question asking for a count of multiple objects for the given attributes. MAC using scaled dot product attention correctly aligned to required objects at different steps. However, the model with biased general* attention focused on the \textit{big things} before noting the condition \textit{left of the green cube}, and failed to filter out irrelevant objects, giving a wrong prediction by aligning to additional objects. In Figure~\ref{fig:mac-analysis-appendix-2}, MAC model using scaled dot product attention focuses on the key objects \textit{purple metal object}, \textit{brown rubber objects}, and \textit{green blocks} in both question and the image, so it can successfully give the correct answer \textit{yes}. However, the model trained with biased general* attention focused on \textit{green blocks} in the question in the last two steps but failed to find the target in the image, thus giving a wrong prediction \textit{no}. These examples clearly align with the finding in Section~\ref{sec:breakdown} that scaled dot product can learn better alignment than biased general* attention for questions querying multiple objects.

\section{Conclusion}
We systematically examined the role of attention alignment score calculation in vision-and-language tasks, including VQA, T\&I Match, and T2I Gen. We found that the scaled dot product can be the best attention alignment calculation for either cross or self-attention in overall VL tasks while the appropriate position of visual and textual information may vary from different VL tasks/datasets. Based on the breakdown analysis, we found out that the type of image objects and their textual description would affect the performance of different attention calculation functions. It is hoped that our analysis provides a great insight into the selection of the most effective attention alignment for different VL tasks.

\newpage
\bibliography{anthology,custom}
\bibliographystyle{acl_natbib}

\appendix


\section{Appendix}\label{sec:qualitative-appendix}

In this section, we demonstrate some more attention alignment examples of best and worst performing attention methods for each task we investigated.

\subsection{Additional Qualitative Examples - VQA}

\begin{figure}[t]
    \centering
    \includegraphics[width=\linewidth]{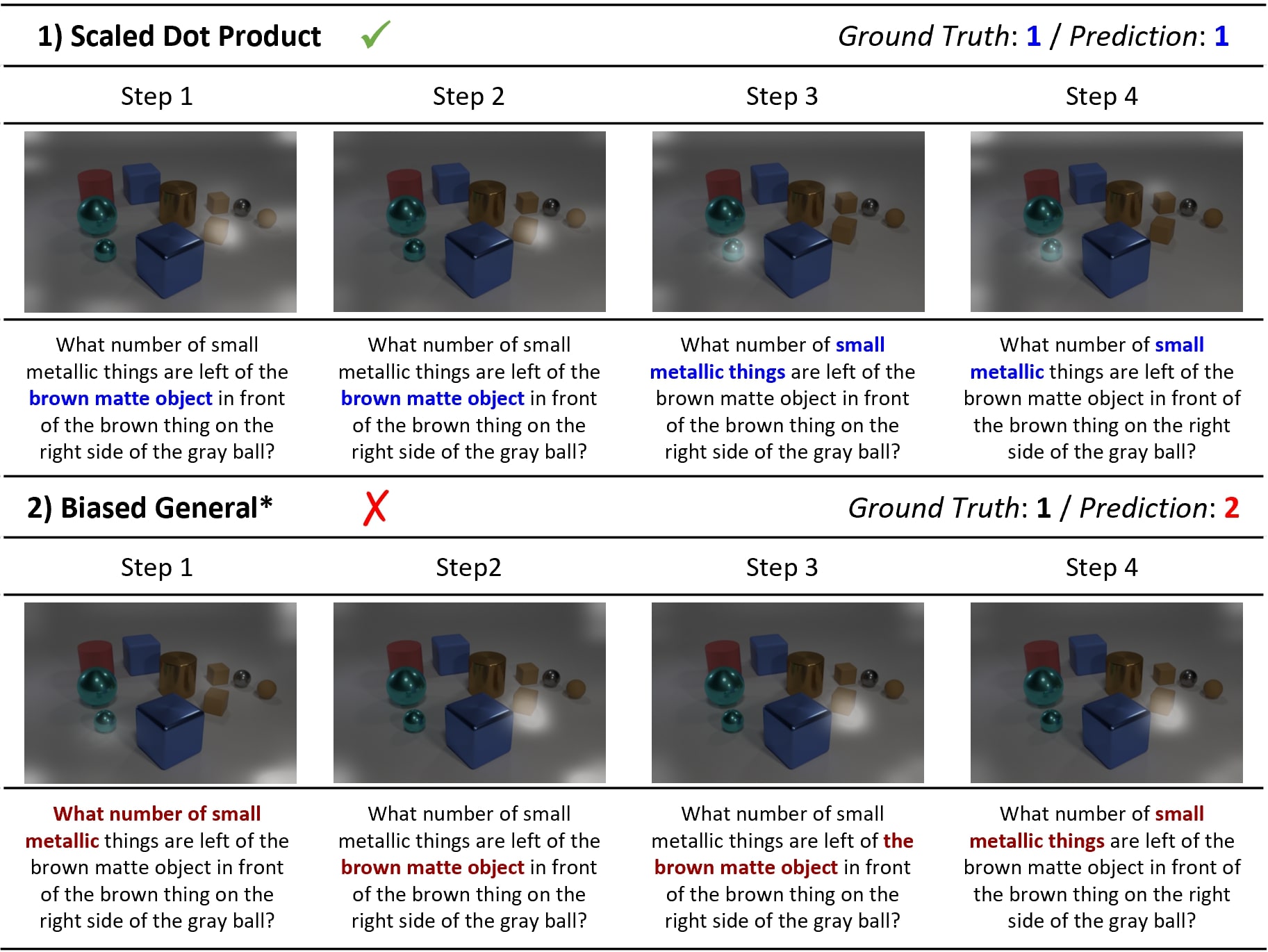}
    \caption{Qualitative examples of VQA-CLEVR from the MAC trained by different attention functions.}
    \label{fig:mac-analysis-appendix}
\end{figure}

We include more comparison examples for MAC model in this section to show the difference between scaled dot product (best) and biased general* (worst) in the VQA context. In \figurename~\ref{fig:mac-analysis-appendix}, a question \textit{what number of small metallic things are left of the brown matte object in front of the brown thing on the right side of the gray ball} is raised towards an image with several cylinders, cubes and spheres. The MAC model with scaled dot product attention is able to correctly focus on the \textit{brown matte object} from both the question and the image, while putting slight attention on \textit{the brown thing on the right side} as mentioned in the question. Then in step 3 and 4 the model is able to locate the \textit{small metallic thing} on the left in the image as guided by the question context, giving a correct prediction of \textit{1}. However, the MAC model trained with biased general* attention slightly focuses on the target metallic object at the very beginning, and shifts its main attention to the \textit{brown matte object} in the consecutive steps, which is not the final target the question is asking for, therefore it fails to make a correct prediction.

\begin{figure}[t!]
    \centering
    \includegraphics[width=\linewidth]{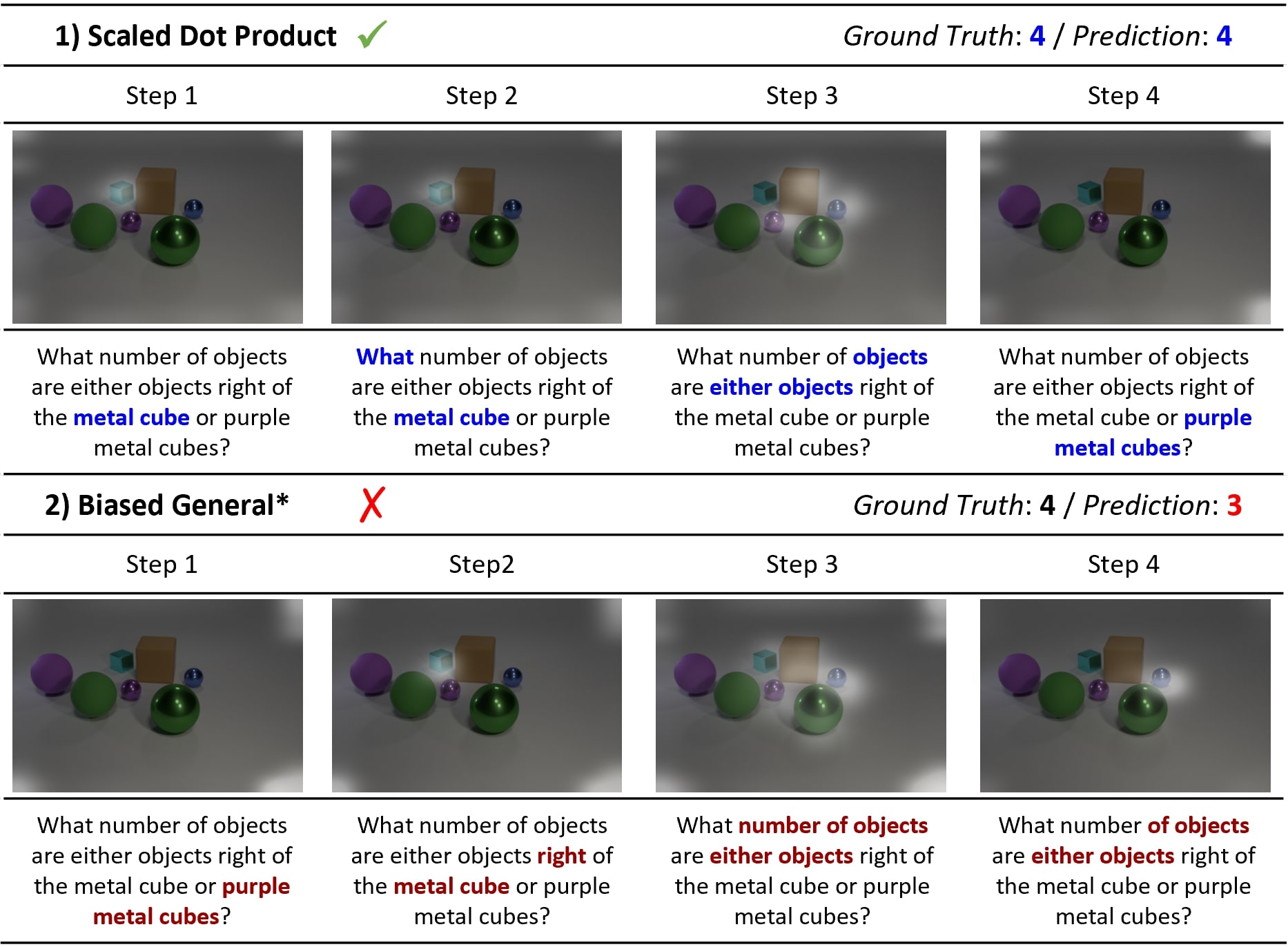}
    \caption{Qualitative examples of VQA-CLEVR from the MAC trained by different attention functions.}
    \label{fig:mac-analysis}
\end{figure}

Figure \ref{fig:mac-analysis} shows a picture featuring several cubes and spheres. With a scaled dot product for attention score calculation, when the model focuses on the keyword \textit{metal cubes} from the textual question, the only metal cube in the image is emphasized during the first two steps. Then, it correctly detects 4 objects from the image by highlighting the keywords \textit{objects} and \textit{either objects}. Additionally, the model looks for the \textit{purple metal cube} from the picture as asked by the question, but it does not exist in the picture so none of the objects are highlighted at step 4. However, the model with the biased general attention tries to count \textit{the number of objects} on the right of the metal cube in step 3 but it inaccurately focuses on the metal cube itself in addition to the correct ones. In the last step the model puts more focus on the farthest right objects, resulting in a wrong prediction of \textit{3}. 

\begin{figure}[t]
    \centering
    \includegraphics[width=\linewidth]{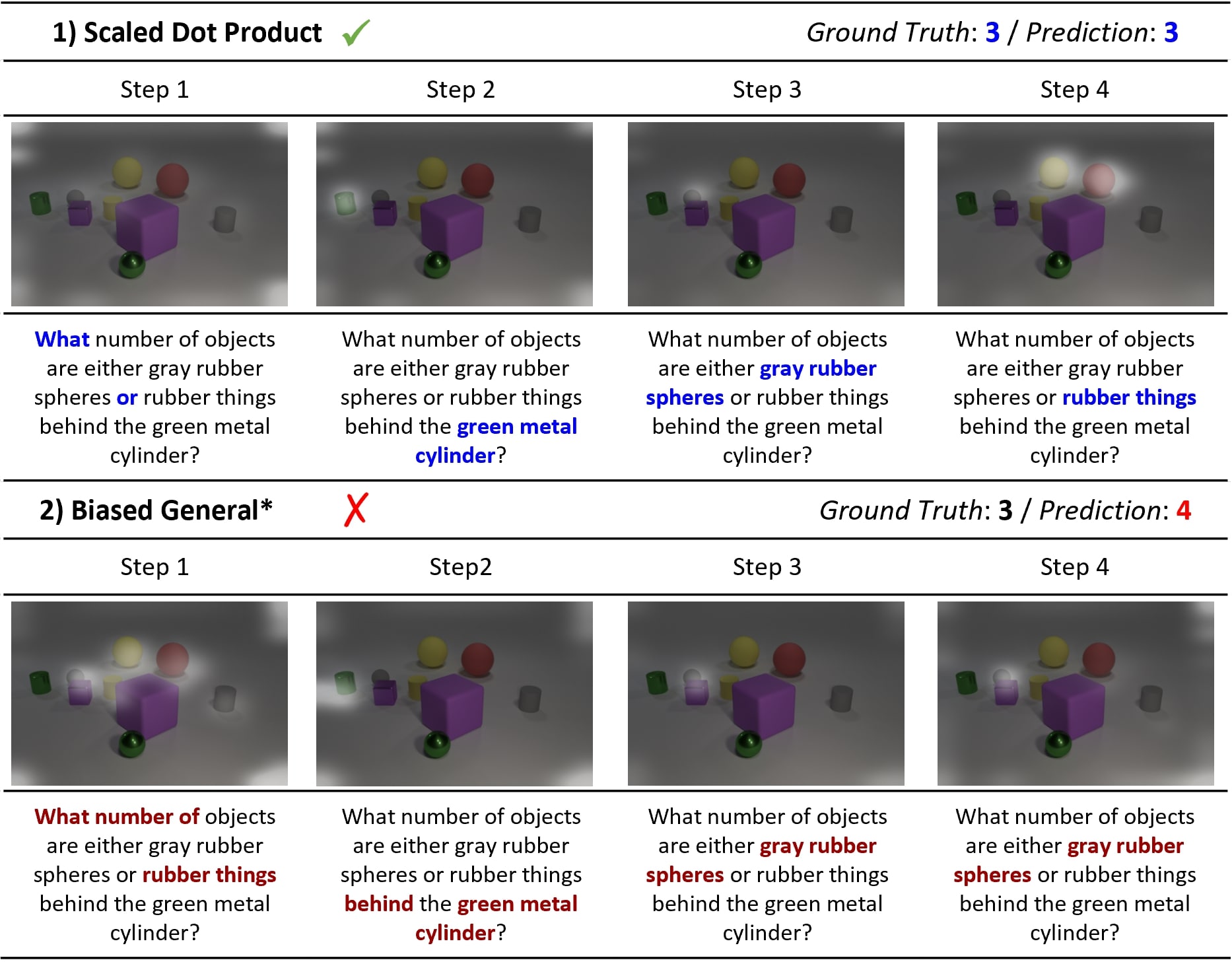}
    \caption{Qualitative examples of VQA-CLEVR from the MAC trained by different attention functions.}
    \label{fig:mac-analysis-appendix-3}
\end{figure}

In Figure~\ref{fig:mac-analysis-appendix-3}, a question \textit{what number of objects are either gray rubber spheres or rubber things behind the green metal cylinder} is asked. MAC model using scaled dot product attention approaches this question by firstly attending to \textit{what} and \textit{or} in the question. Then it focuses on the relevant objects \textit{green metal cylinder}, \textit{gray rubber spheres}, and remaining \textit{rubber things} in both question and the image, so it can successfully give the correct answer \textit{3}. However, the model trained with biased general* attention firstly focused on the \textit{number of rubber things} before noting the condition \textit{behind the green metal cylinder}, so it failed to filter out irrelevant objects, giving a wrong prediction \textit{4}.

\begin{figure}[t!]
    \centering
    \includegraphics[width=\linewidth]{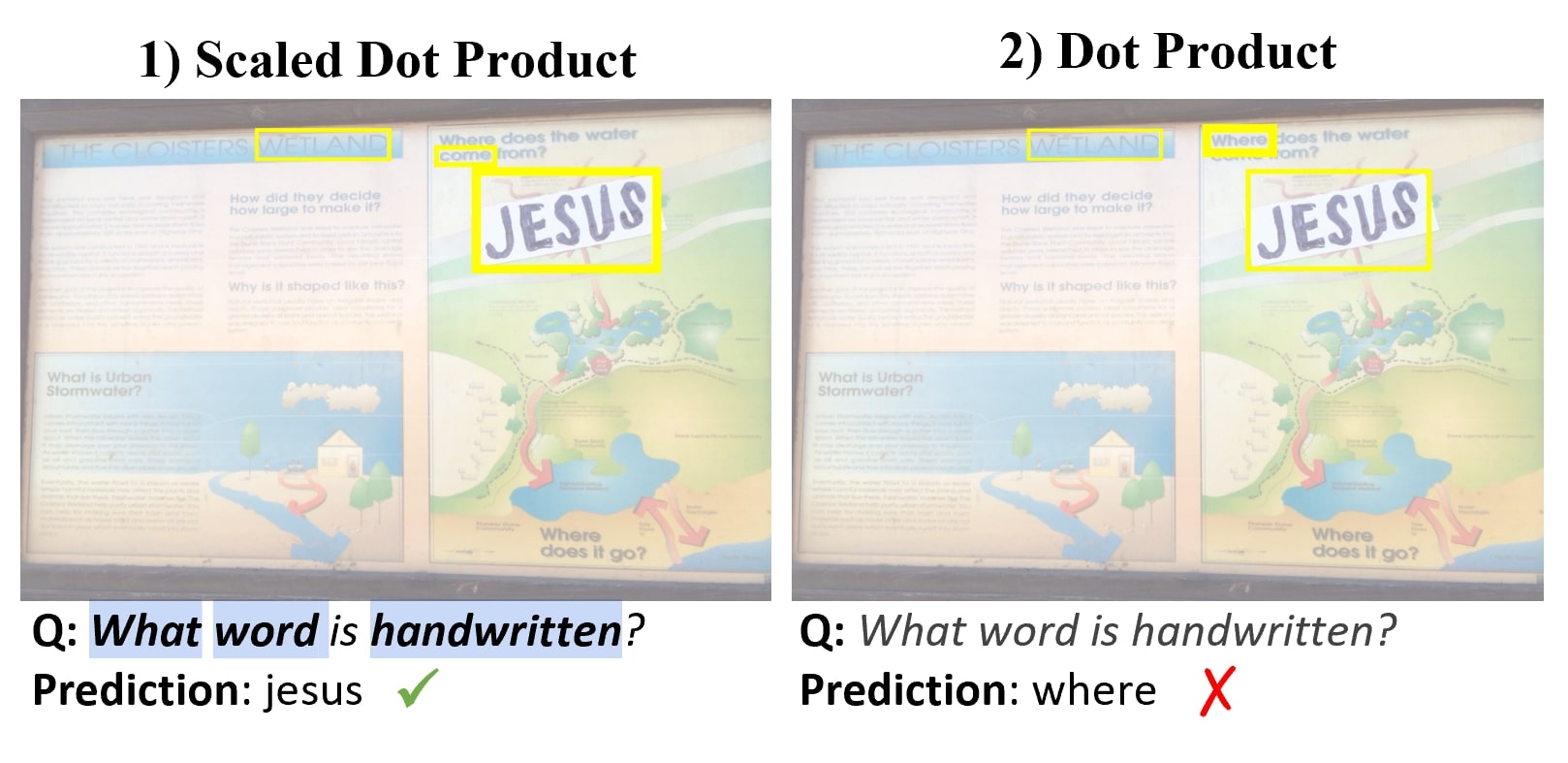}
    \caption{Qualitative examples of TVQA-TextVQA from the M4C trained by different attention functions. Question words that receive attention weights greater than 0.01 are indicated in bold and coloured in blue.}
    \label{fig:m4c-analysis}
\end{figure}

\begin{figure}[t!]
    \centering
    \includegraphics[width=\linewidth]{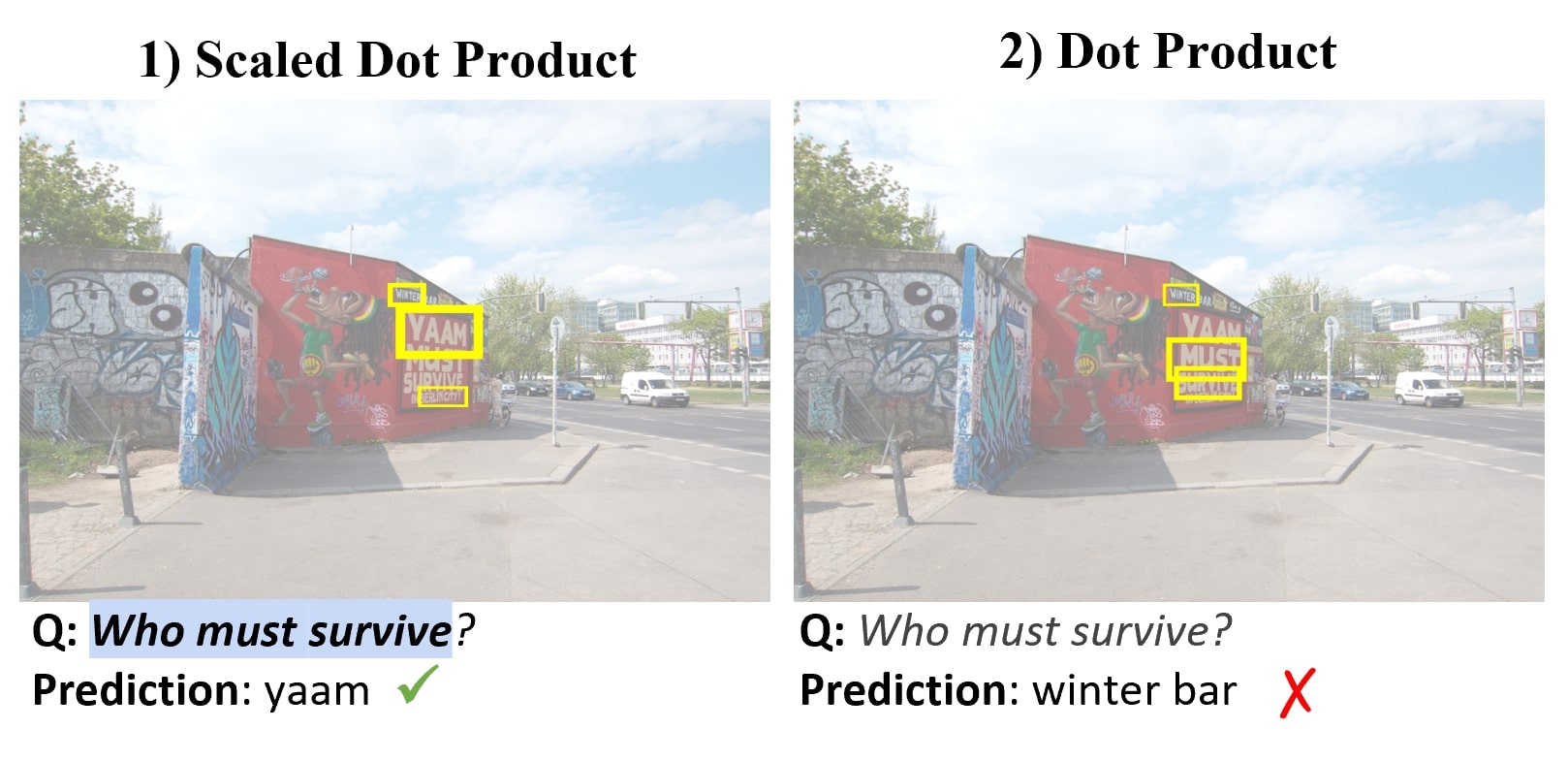}
    \caption{Qualitative examples of TVQA-TextVQA from the M4C trained by different attention functions. Question words that receive attention weights greater than 0.01 are indicated in bold and coloured in blue.}
    \label{fig:m4c-analysis-appendix-1}
\end{figure}

\subsection{Additional Qualitative Examples - TVQA}

We include some qualitative examples for M4C model in this section to show the difference between scaled dot product (best) and dot product attention (worst) in the context of TVQA.

Figure \ref{fig:m4c-analysis} shows that the model with scaled dot product focused on the keywords \textit{what}, \textit{word} and \textit{handwritten} to focus on the handwritten word \textit{jesus} in the image and retrieved the correct OCR token with highest attention weight. However, with dot product attention, all the question words received little attention by the model ($< 0.01$), failing to find the appropriate OCR token in the image.  

In Figure~\ref{fig:m4c-analysis-appendix-1}, all the three words from the question \textit{who must survive} received attention $> 0.01$ in the scaled dot product model, and the target OCR answer in the image received top attention among all OCR tokens. However, the dot product model put much less attention on all question words, instead the OCR tokens for \textit{must} and \textit{survive} in the image were receiving top attention weights, followed by OCR token \textit{winter} which is irrelevant to the question. Therefore scaled dot product model predicted correctly but dot product model did not.

\begin{figure}[t!]
    \centering
    \includegraphics[width=\linewidth]{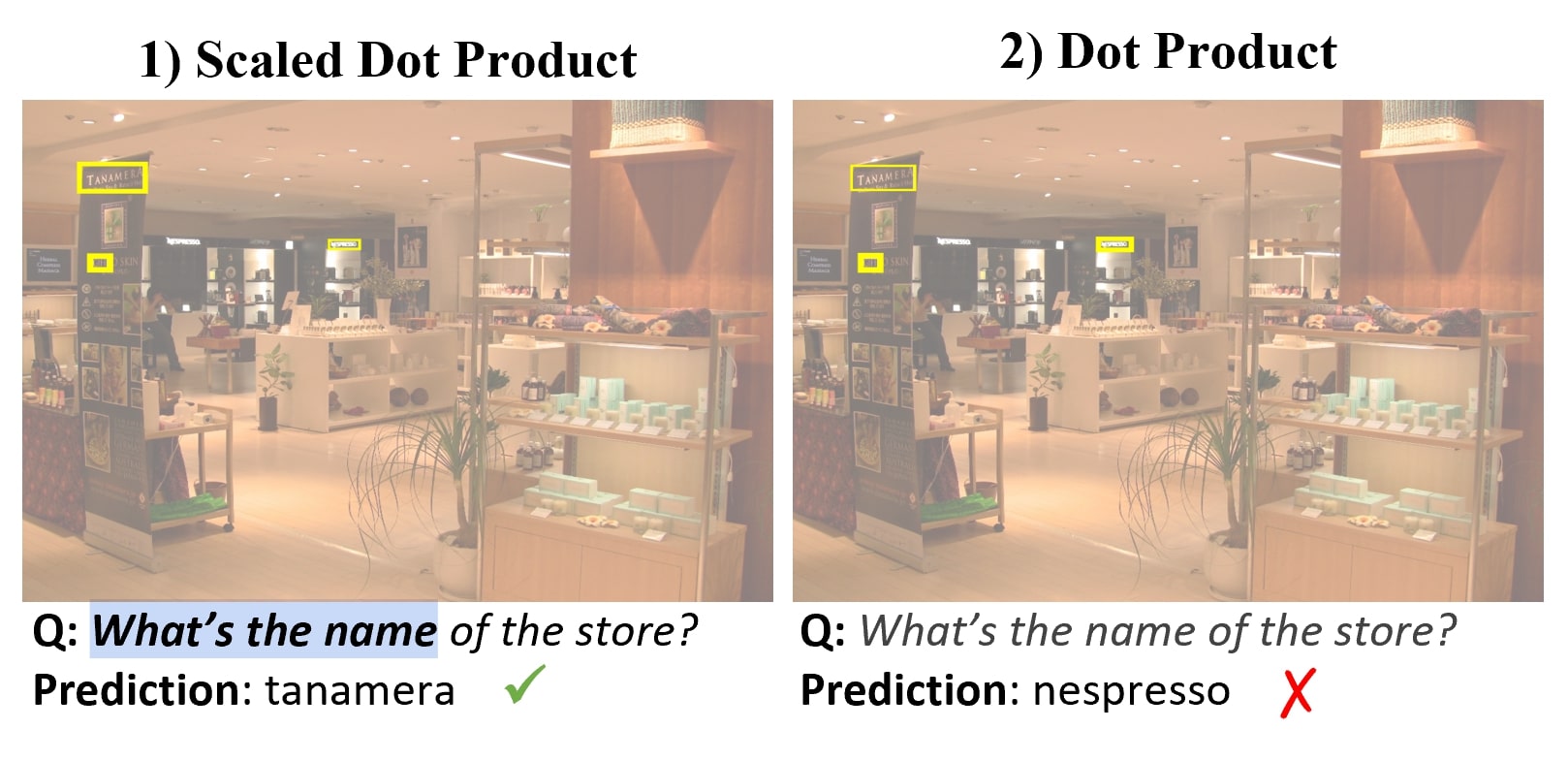}
    \caption{Qualitative examples of TVQA-TextVQA from the M4C trained by different attention functions. Question words that receive attention weights greater than 0.001 are indicated in bold and coloured in blue.}
    \label{fig:m4c-analysis-appendix-2}
\end{figure}

\begin{figure}[t!]
    \centering
    \includegraphics[width=\linewidth]{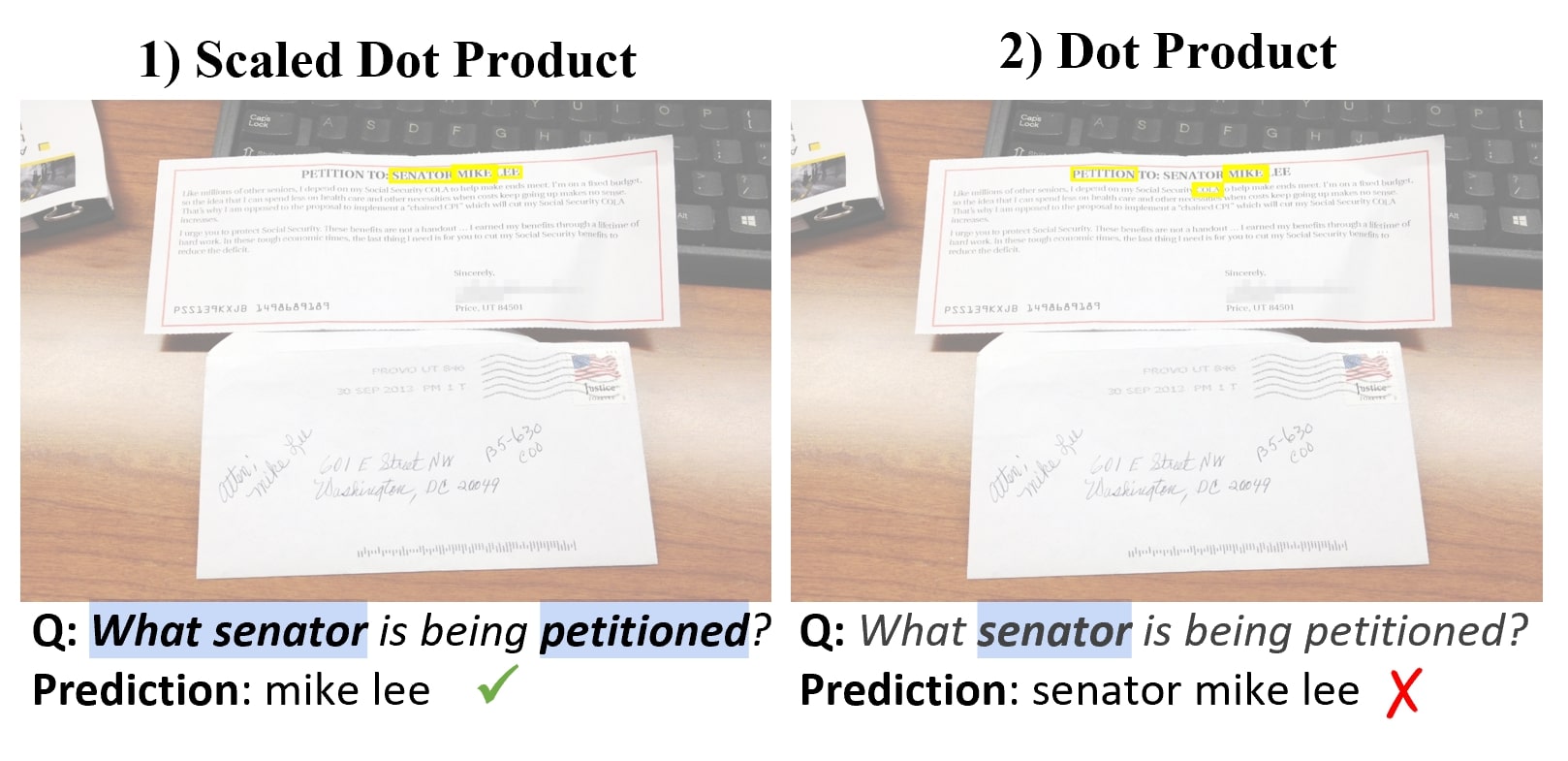}
    \caption{Qualitative examples of TVQA-TextVQA from the M4C trained by different attention functions. Question words that receive attention weights greater than 0.001 are indicated in bold and coloured in blue.}
    \label{fig:m4c-analysis-appendix-3}
\end{figure}

In Figure~\ref{fig:m4c-analysis-appendix-2}, keywords \textit{what's the name} from the question \textit{what's the name of the store} received attention $> 0.001$ in the scaled dot product model, similarly the dot product model put much less attention on all question words, and none of the question words received attention $> 0.001$. Both models put most attention weights on the OCR token \textit{gift} from the image, but scaled dot product managed to put more focus on the store name \textit{tanamera} than the coffee brand name \textit{nespresso}, which is the opposite case of the dot product model. Therefore scaled dot product model predicted correctly but dot product model did not.

In the example shown by Figure~\ref{fig:m4c-analysis-appendix-3}, there are lots of OCR tokens present in the image, making it more difficult to retrieve the correct answer tokens. As we can see from the picture, the model learned using dot product attention diverted its top attention to unrelated OCR token \textit{cola}, and the top 3 OCR tokens receiving highest attention (\textit{mike}, \textit{cola} and \textit{petition}) are not aligned with the predicted answer tokens (\textit{senator mike lee}), while the model learned using scaled dot product attention put highest attention to expected or related OCR tokens that are aligned with the ground truth answers.

\subsection{Additional Qualitative Examples - T\&I Match}

\begin{figure}[t!]
    \centering
    \includegraphics[width=\linewidth]{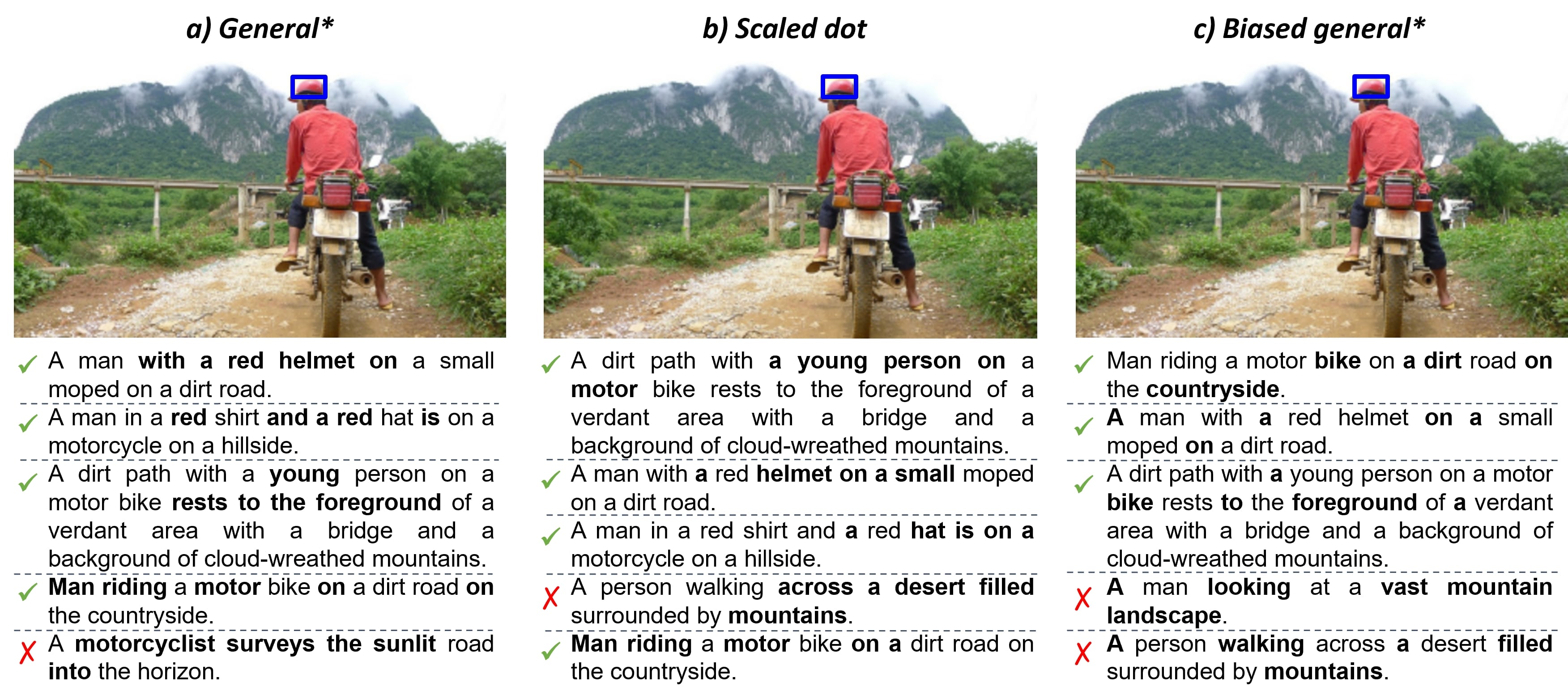}
    \caption{Qualitative examples of T\&I Match-MSCOCO with SCAN by different attention functions.}
    \label{fig:scan-analysis}
\end{figure}

\begin{figure}[t]
    \centering
    \includegraphics[width=\linewidth]{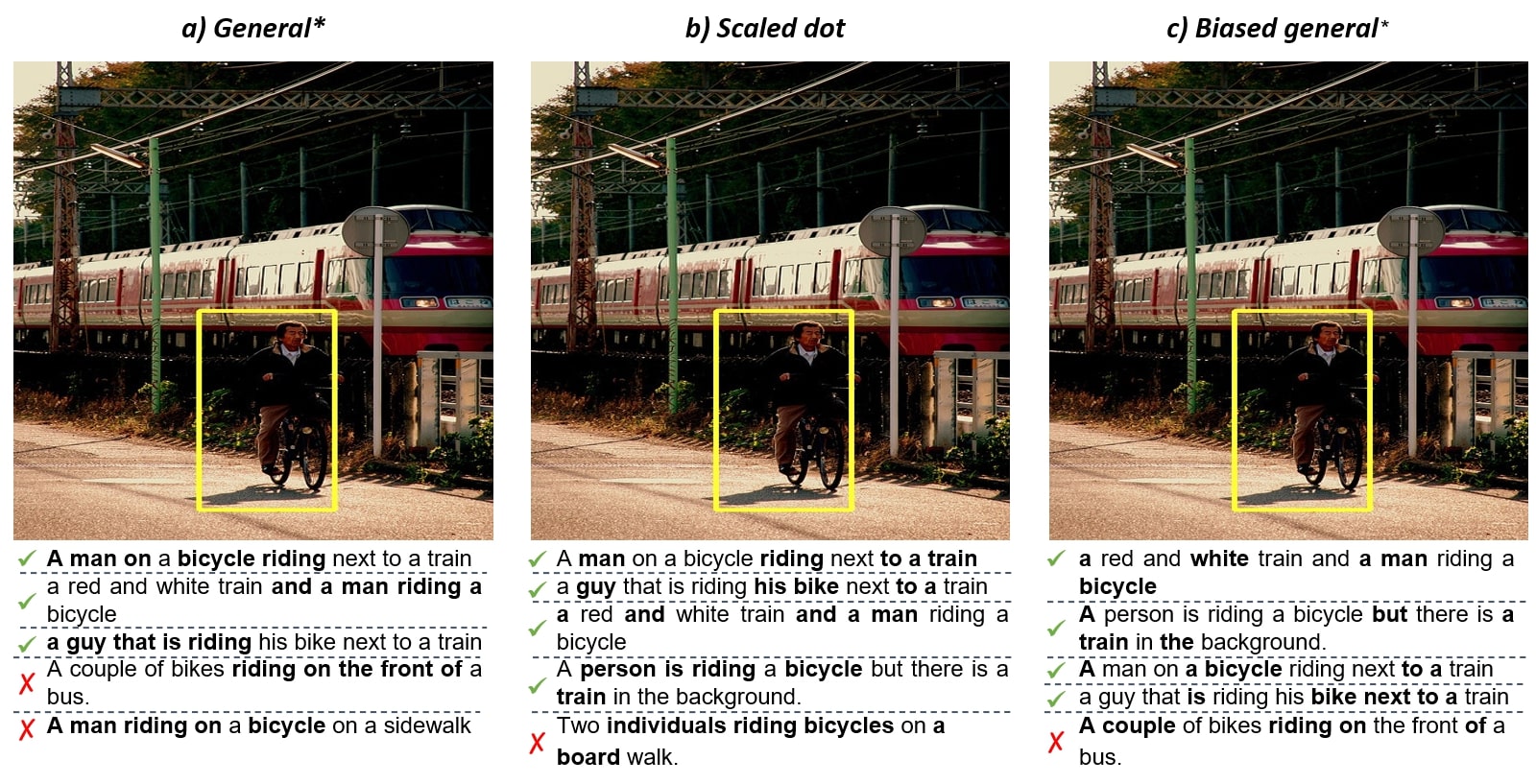}
    \caption{Qualitative examples of T\&I Match-MSCOCO from the SCAN trained by different attention functions.}
    \label{fig:scan-analysis-appendix-1}
\end{figure}

\begin{figure}[t]
    \centering
    \includegraphics[width=\linewidth]{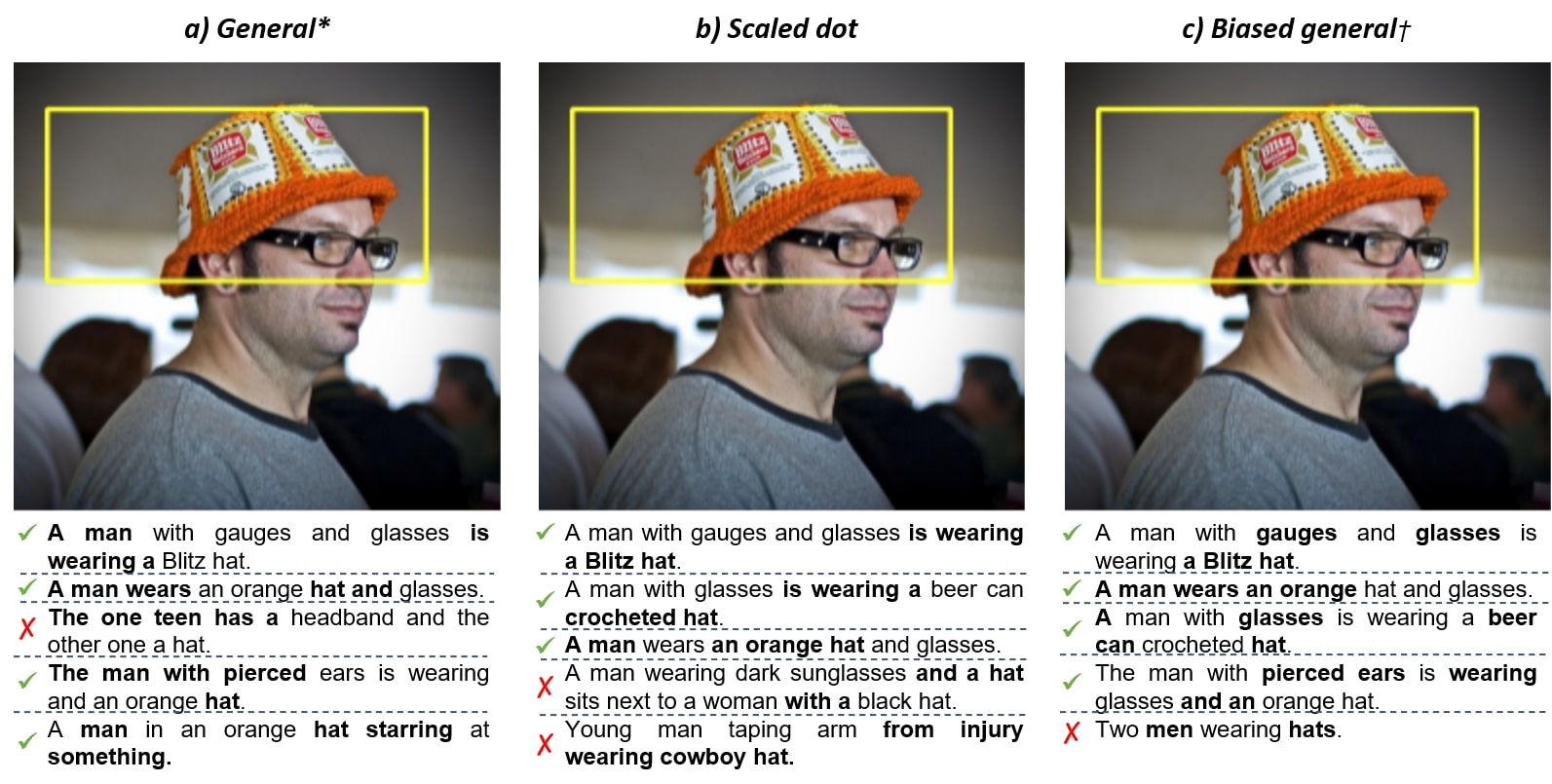}
    \caption{Qualitative examples of T\&I Match-Flickr30k from the SCAN trained by different attention functions.}
    \label{fig:scan-analysis-appendix-2}
\end{figure}

\begin{figure}[t]
    \centering
    \includegraphics[width=\linewidth]{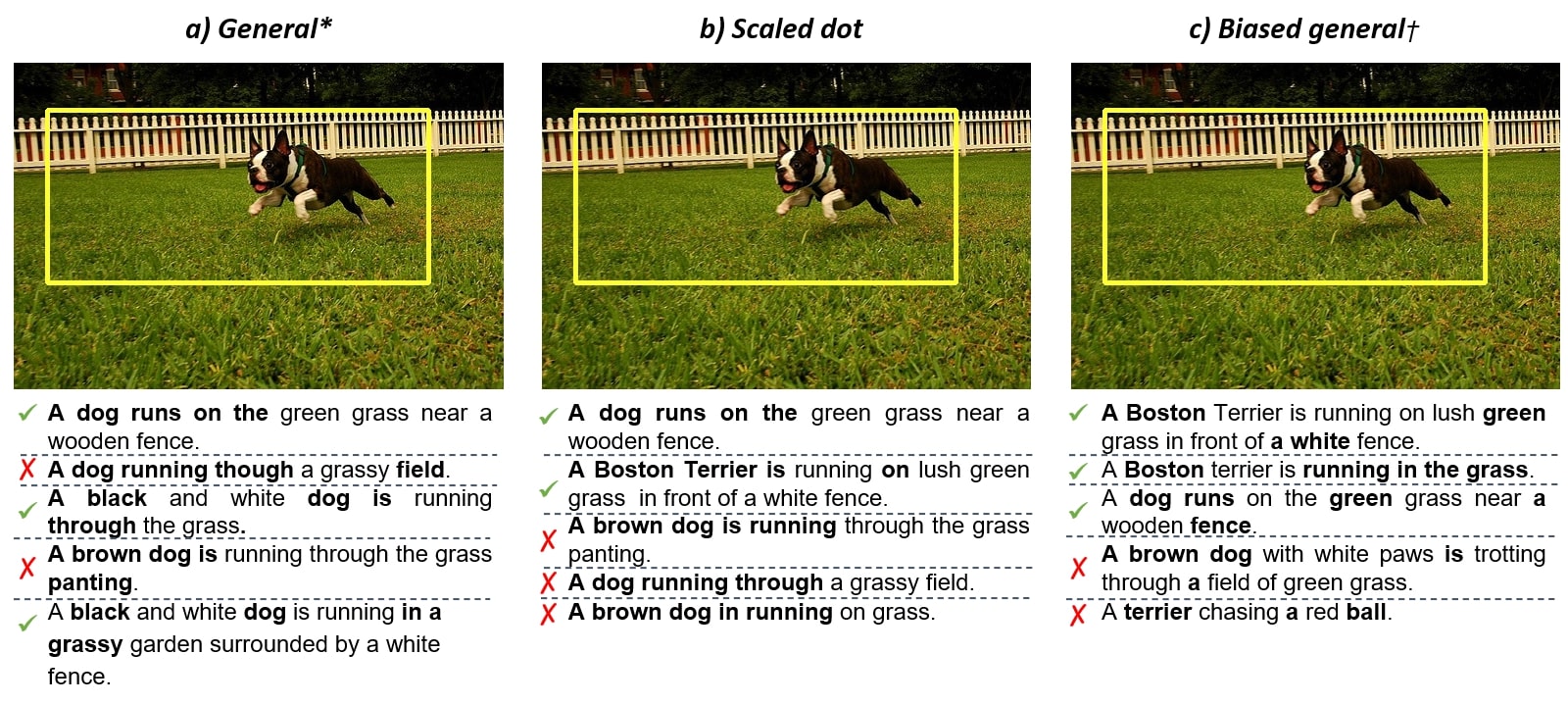}
    \caption{Qualitative examples of T\&I Match-Flickr30k from the SCAN trained by different alignment functions.}
    \label{fig:scan-analysis-appendix-3}
\end{figure}

In this section, we visualize some examples for T\&I Match models that show the attention received by retrieved captions with respect to the selected object region. Figure \ref{fig:scan-analysis} shows some examples for T\&I Match. With general attention* and scaled dot product, the SCAN model is able to include 4 correct captions among the top 5 retrieved captions, while biased general attention can only extract 3 correct captions. With respect to the helmet indicated by the blue box in the image, both general attention and scaled dot product can put more focus on the words \textit{red}/\textit{helmet}/\textit{hat} from the caption, whereas the biased general attention would rather focus on prepositions, determinants or other objects.

In Figure~\ref{fig:scan-analysis-appendix-1} we can see that the best two models (i.e. models trained with general* attention or scaled dot product attention) can capture all key elements, \textit{man, bicycle/bike, riding}, as the top attended words from the retrieved captions most of the time. However, the model trained using biased general* attention would capture at most one key element from each retrieved caption, and pay high attention to preposition, determinants or words related to other object regions. In the example shown by Figure~\ref{fig:scan-analysis-appendix-2}, the model trained with scaled dot product attention can always capture the main object \textit{hat} from all the retrieved captions, while the other two models sometimes fail to do so. In Figure~\ref{fig:scan-analysis-appendix-3}, all three models sometimes wrongly recognise the dog's color (i.e. \textit{brown dog} in wrongly retrieved captions). However, the best two models can retrieve the caption that is not in the ground truth list but also semantically matched to the given image (i.e. \textit{A dog running through a grassy field}). The worst model trained with biased general$\dagger$ attention fails to do so, and it sometimes attends to objects from the caption that is not actually in the image (e.g. \textit{red ball}). 

\subsection{Additional Qualitative Examples - T2I Gen}
In this section, we visualize and compare the best and the worst T2I Gen model.

\begin{figure}[t!]
    \centering
    \includegraphics[width=\linewidth]{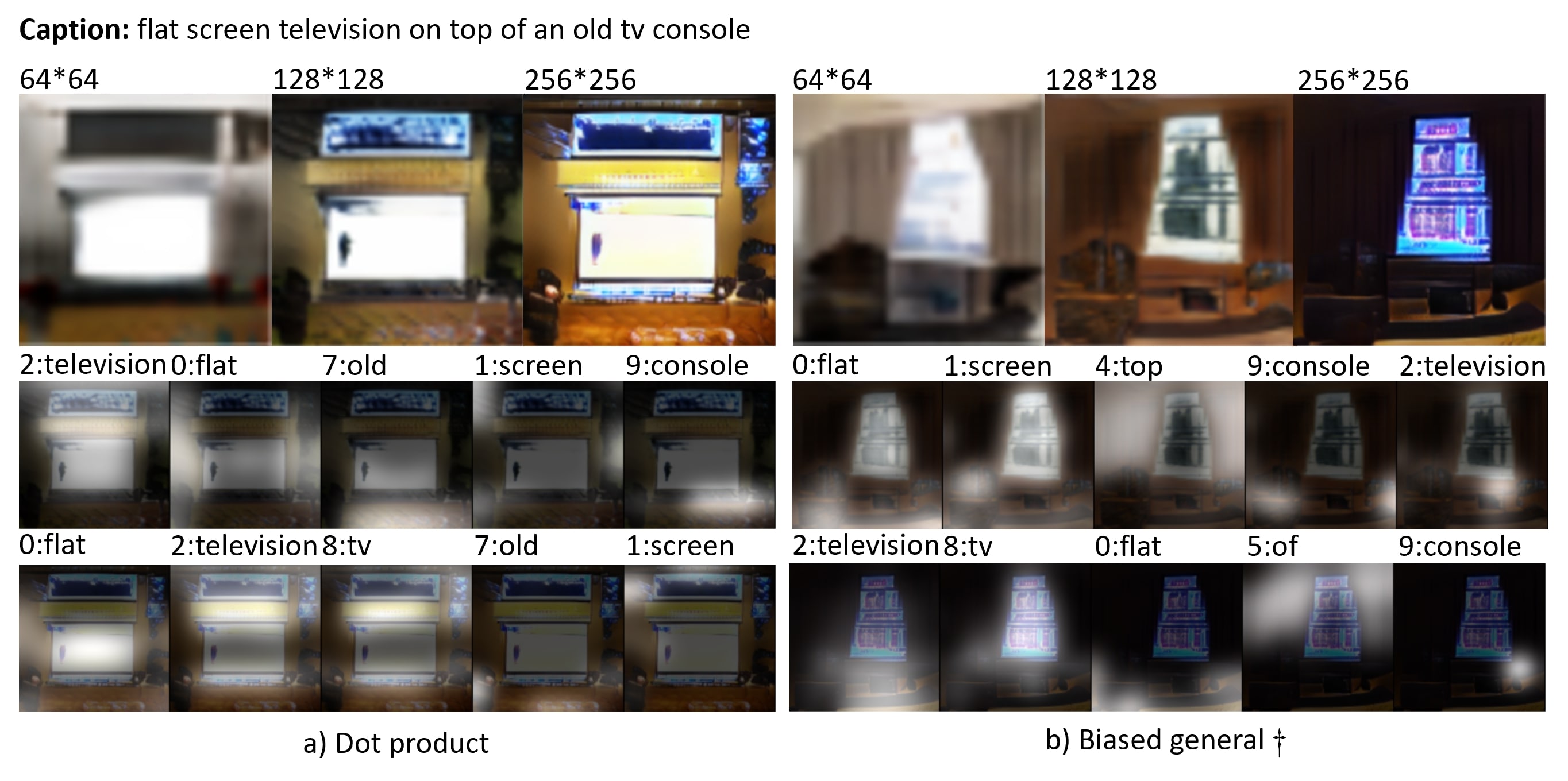}
    \caption{Qualitative examples of T2I Gen-MSCOCO with AttnGAN trained by different attention functions.}
    \label{fig:attngan-analysis}
\end{figure}

Figure \ref{fig:attngan-analysis} shows the images generated by AttnGAN using both the best and the worst attention, dot product and biased general$\dagger$ respectively. AttnGAN with a dot product, can generate a relatively more realistic image. From a low resolution picture, the model focuses on the words based on the following order, \textit{television}, \textit{flat}, \textit{old}, \textit{screen}, \textit{console}, in order to refine the image to include the objects and corresponding features gradually. Compared to that, the biased general attention model generates a surrealistic image by focusing on \textit{flat}, \textit{screen}, \textit{top}, \textit{console}, \textit{television} in the first step.

\begin{figure}[t!]
    \centering
    \includegraphics[width=\linewidth]{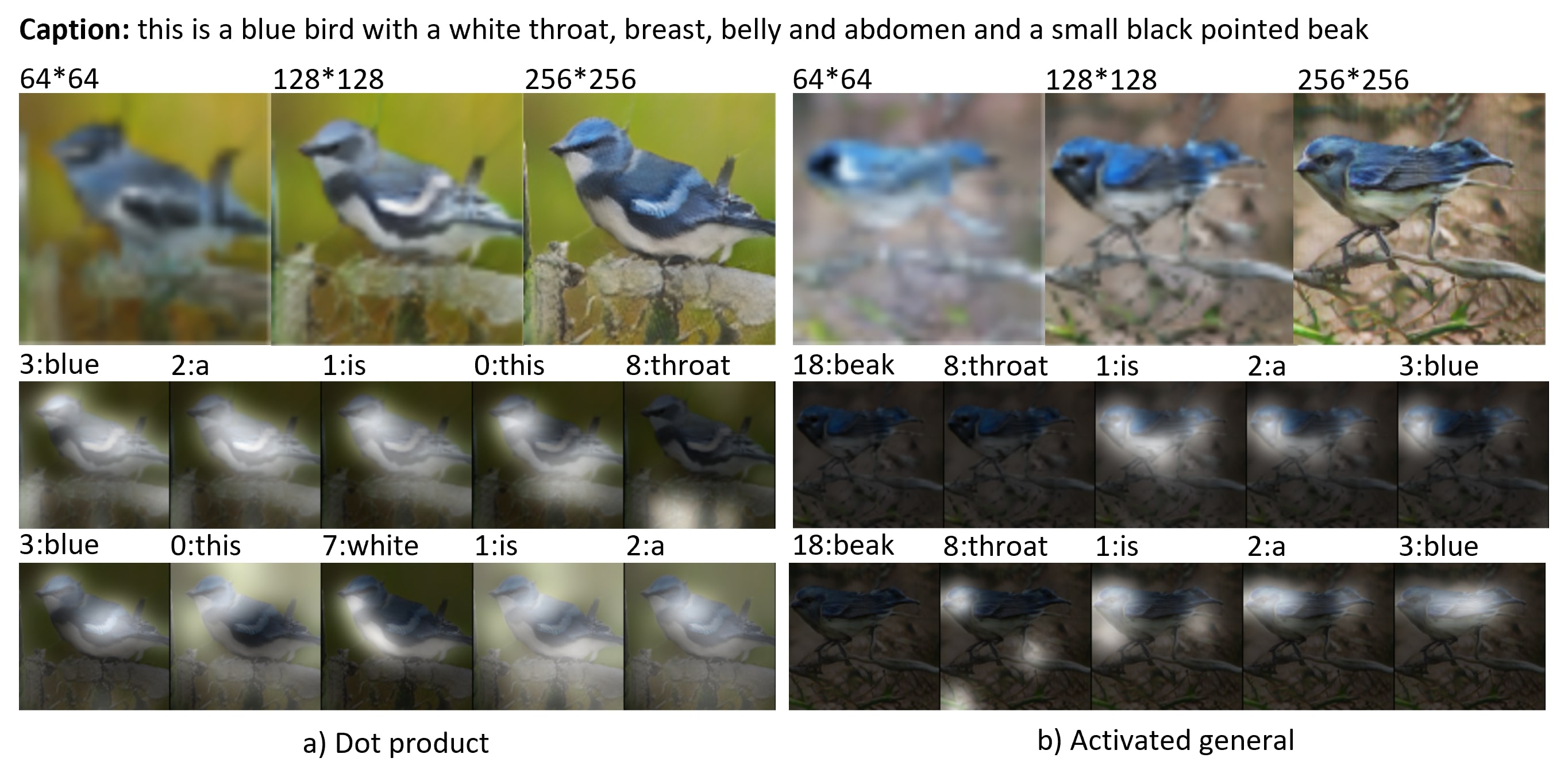}
    \caption{Qualitative examples of T2I Gen-CUB from the AttnGAN trained by different attention functions.}
    \label{fig:attngan-analysis-appendix}
\end{figure}
\begin{figure}[t!]
    \centering
    \includegraphics[width=\linewidth]{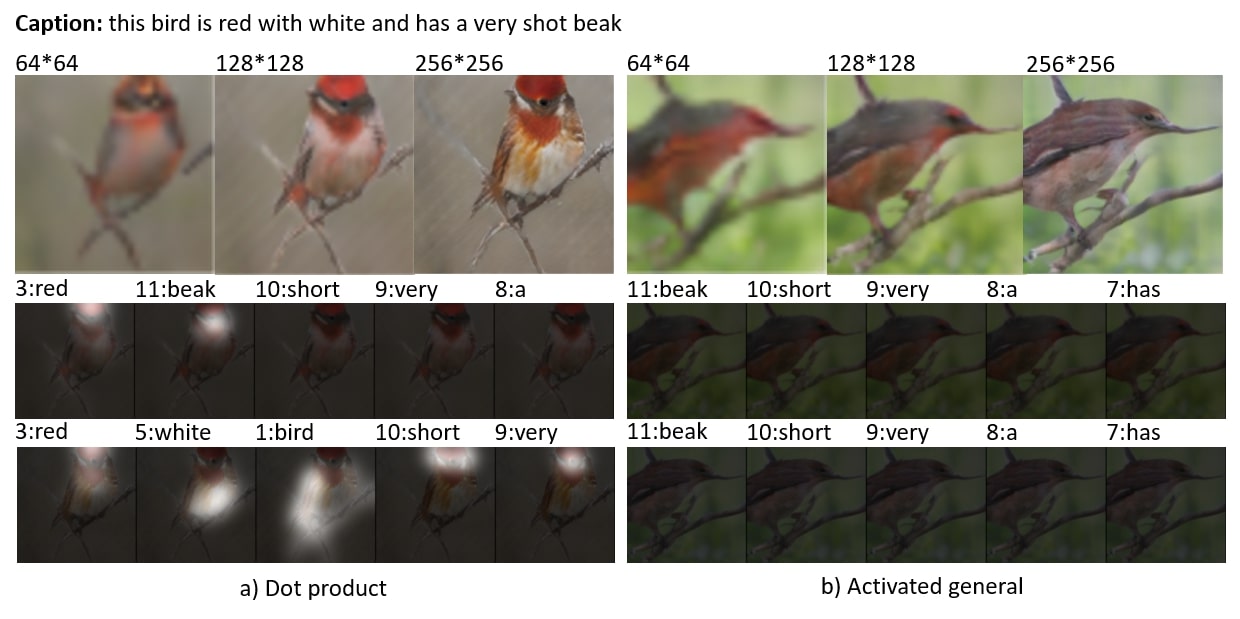}
    \caption{Qualitative examples of T2I Gen-CUB from the AttnGAN trained by different attention functions.}
    \label{fig:attngan-analysis-appendix-2}
\end{figure}

\begin{figure}[t!]
    \centering
    \includegraphics[width=\linewidth]{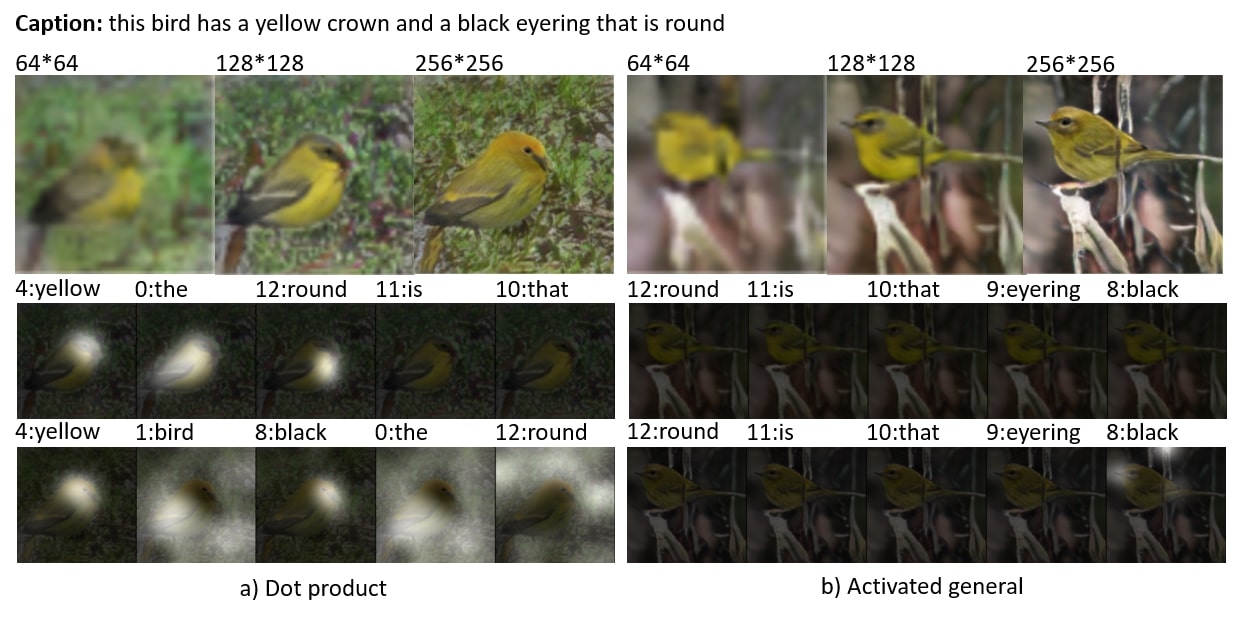}
    \caption{Qualitative examples of T2I Gen-CUB from the AttnGAN trained by different attention functions.}
    \label{fig:attngan-analysis-appendix-3}
\end{figure}
In Figure~\ref{fig:attngan-analysis-appendix}, the images generated by two models are highly similar but the worst model trained with activated general attention fails to attend to the key word \textit{white}, so the bird it generated in the picture does not clearly have a white throat and chest. 

In Figure~\ref{fig:attngan-analysis-appendix-2} and Figure~\ref{fig:attngan-analysis-appendix-3}, the activation function used in the attention mechanism of the worst model makes it difficult to differentiate among the caption words when their attention weights are all very low. Therefore the model fails to attend to any useful facts in each attention layer, which makes it impossible to provide interpretability of model decision, despite generating an image that can roughly match the description in Figure~\ref{fig:attngan-analysis-appendix-3}. In Figure~\ref{fig:attngan-analysis-appendix-2} the quality of the generated image is even worse - the feature of the bird does not match with the key phrases in the description (i.e. \textit{red with white, short beak}).

Overall, based on the qualitative analysis in different VL tasks, we reveal that the better attention alignment calculation function can produce better interpretability in terms of the prediction. 

\end{document}